\documentclass{article}
\usepackage{iclr2026_conference,times}
\iclrfinalcopy

\usepackage{amsmath,amsfonts,bm}

\def\eqref#1{equation~\ref{#1}}

\def\1{\bm{1}}

\DeclareMathAlphabet{\mathsfit}{\encodingdefault}{\sfdefault}{m}{sl}
\SetMathAlphabet{\mathsfit}{bold}{\encodingdefault}{\sfdefault}{bx}{n}

\usepackage{hyperref}
\usepackage{url}
\usepackage{graphicx}
\usepackage{amsmath,amssymb,bm}
\usepackage{booktabs}
\usepackage{multirow}
\usepackage{xcolor}
\usepackage{tikz}
\usetikzlibrary{positioning}
\graphicspath{{figures/}}

\newcommand{\state}{\boldsymbol{q}}
\newcommand{\feat}{\boldsymbol{\Phi}}
\newcommand{\rawfeat}{\widetilde{\boldsymbol{\Phi}}}   %
\newcommand{\lift}{\boldsymbol{\Psi}}
\newcommand{\prop}{\boldsymbol{G}}
\newcommand{\Lin}{\boldsymbol{L}}
\newcommand{\Zs}{\boldsymbol{Z}}
\newcommand{\Zd}{\boldsymbol{Z}^\dagger}
\newcommand{\qL}{\state^\dagger_{\mathrm L}}

\title{SALTO: Solver-Anchored Lifted Transfer Operators for Nonlinear Event Inference}

\author{Qi Wang\\
Department of Aerospace Engineering\\
San Diego State University\\
San Diego, CA 92182, USA\\
\texttt{qwang4@sdsu.edu}
\And
Rylie Foster\\
Computational Science Research Center\\
San Diego State University\\
San Diego, CA 92182, USA\\
Department of Earth System Science\\
University of California, Irvine, CA 92697, USA}

\begin{document}
\maketitle

\begin{abstract}
Learned dynamical models can accurately predict trajectories while producing unreliable sensitivities when differentiated or transposed, limiting their use in inverse problems and scientific decision-making. Classical adjoints provide exact local derivatives, but a single adjoint cannot represent the nonlinear response of a system to a family of finite-amplitude perturbations. We introduce \textsc{SALTO} (\emph{Solver-Anchored Lifted Transfer Operators}), a Koopman-based framework for inferring finite-amplitude sources. In the lifted space, the nonlinear dynamics are approximated by a linear propagator, and its transpose carries each sensor backward in a single solve. The result is a reusable lifted adjoint that predicts the measurement produced by any candidate localized event, such as a source of given position and amplitude, with one encoder pass and an inner product, so that the search needs no repeated nonlinear simulation. To give the model the simulator's exact local behavior, \textsc{SALTO} imposes three structural constraints: it anchors the physical-state dynamics to the simulator's linearized propagator, centers the learned observables so that they vanish to second order at the reference state, and uses a block-triangular lifted propagator. These constraints guarantee that the model reduces to the simulator's tangent dynamics and classical adjoint as the amplitude vanishes, while the learned observables carry only the finite-amplitude correction. On three nonlinear PDE systems, \textsc{SALTO} predicts the measurements of finite-amplitude sources to within $2$--$8\%$ where the classical adjoint is in error by $41$--$165\%$, recovers source position and intensity from sparse sensors nearly as accurately as inversion through the solver, and samples the joint posterior of two interacting sources, matching the solver's posterior on reaction--diffusion at a twentieth of its cost. One trained model serves new sensor layouts and measurement times with a single backward solve, and its exact local structure keeps both measurements and their gradients correct as the amplitude vanishes, where unconstrained Koopman models do not. Enforcing exact local structure thus turns a learned lift into a reliable instrument for nonlinear sensitivity and inference.
\end{abstract}

\section{Introduction}

Many problems in the physical sciences ask us to identify the cause of an observed signal. When a contaminant is detected at a downstream sensor, or a disturbance is recorded in a flow, we want to know which event produced it, where and when it occurred, and how strong it was. Such source localization problems arise in environmental monitoring, leak detection and flow diagnosis, and they all depend on how sensitive a measurement is to a perturbation of the state. The standard tool for computing this sensitivity is the adjoint of the tangent linear model. A single backward solve gives the derivative of a measurement with respect to the state at every earlier time, which is why adjoints underlie source localization, data assimilation, sensor placement and gradient-based design \citep{errico1997adjoint, giles2000introduction, Luchini2014adjoint, zaki2025arfm}. The adjoint, however, has a structural limitation. Because it linearizes the dynamics about a base trajectory, the measurement it predicts for a source is linear in the amplitude of that source. For a source of finite amplitude the measurement is a nonlinear functional of the source, and no single adjoint, whether about the base state or about any one nonlinear trajectory, can represent it. Inference therefore needs a sensitivity defined over a whole family of sources rather than a single vector.

\begin{figure}[t]
\centering\includegraphics[width=0.96\textwidth]{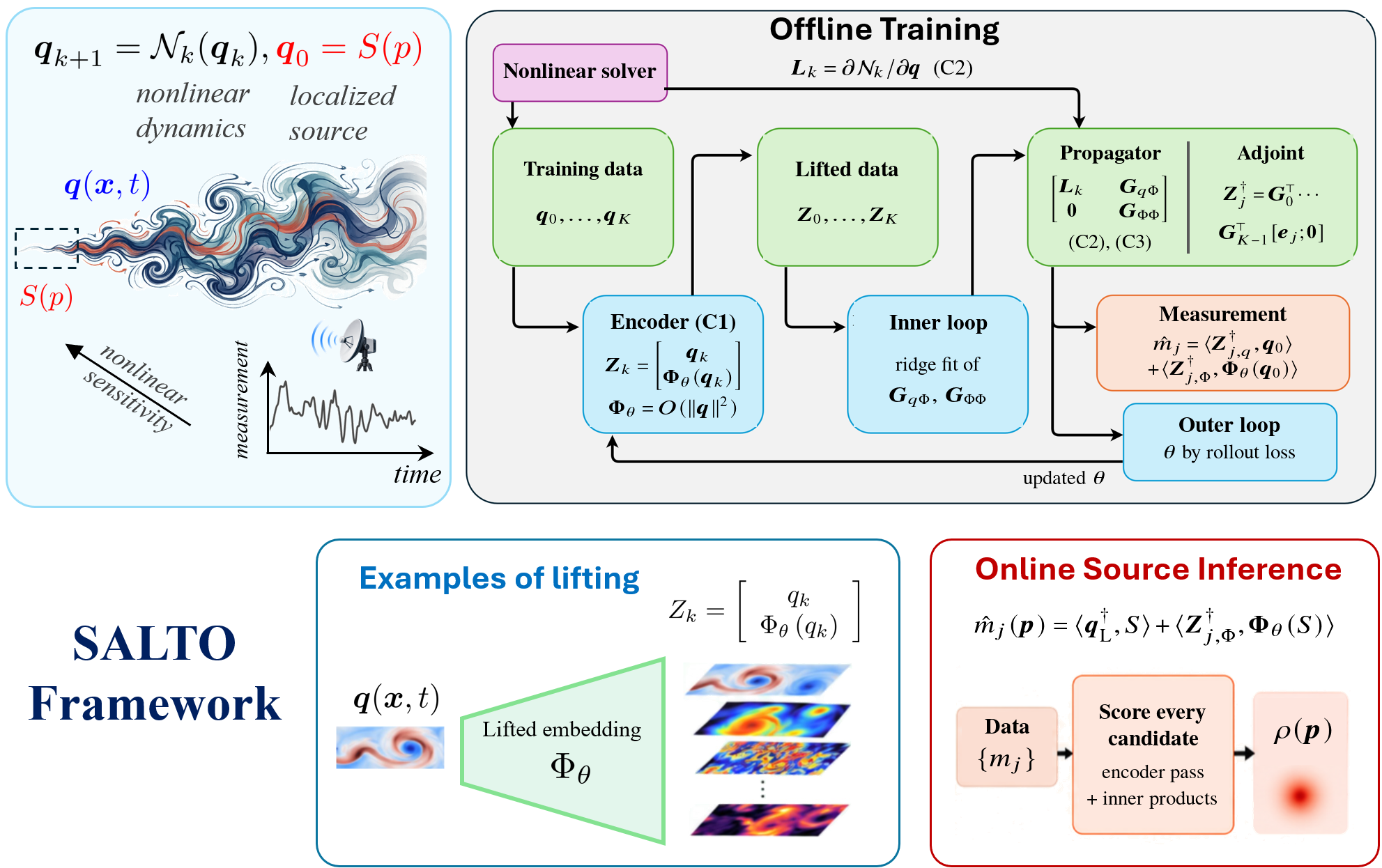}
\caption{\textsc{SALTO} in one picture. Localized sources $S(\boldsymbol p)$ seed a nonlinear flow observed by a few probes. Offline, the solver supplies $\Lin_k$ and the data supply only the finite-amplitude correction, so the lifted model is exact at small amplitude. Online, one backward solve per sensor gives the lifted adjoint $\Zd_j$, and every candidate source is scored by an encoder pass and inner products.}
\label{fig:framework}
\end{figure}

This paper obtains that functional from a single linear adjoint solve by means of a Koopman lift \citep{koopman1931hamiltonian, mezic2005spectral, williams2015data, brunton2022modern}. We augment the state with a set of learned nonlinear observables, so that the augmented, or lifted, state evolves linearly under an approximate Koopman operator. A measurement at the final time is then a linear functional of the lifted state, and carrying it one step backward in time amounts to applying the transpose of that operator. Repeating this back to the time of the source gives what we call the \emph{lifted adjoint} (Fig.~\ref{fig:framework}). Because the lifted adjoint does not depend on the state, it is computed once for each sensor and reused for every candidate source. The measurement predicted for a candidate source is then an inner product between the lifted adjoint and the lifted source. This inner product is a nonlinear functional of the source that costs a single encoder pass to evaluate, and it is exactly what finite-amplitude source inference needs.

The difficulty is that forward accuracy does not make a learned lift a reliable sensitivity. An unconstrained lift must learn the linear dynamics along with the nonlinear ones, even though the solver already provides them exactly, and any error in this linear part enters every sensitivity the lift predicts, including those of vanishingly weak sources. Such errors barely affect trajectory predictions at the training amplitudes, which is why a lift can forecast well and still transpose poorly. We therefore take the linear part from the solver and learn only the nonlinear correction, and we enforce this by construction with three constraints. The learned observables are centered so that they vanish to second order at the base state (C1), the lifted propagator uses the solver's linearized propagator for the physical state (C2), and it is block-triangular so that the learned observables evolve on their own (C3). The lifted model then reduces exactly to the solver's tangent dynamics and the classical adjoint as the amplitude vanishes, and the learned observables carry only the finite-amplitude correction.

\paragraph{Contributions.} (i) \textbf{A solver-anchored lifted adjoint that is exact at small amplitude.} We formulate finite-amplitude sensitivity with a Koopman lift whose state-independent adjoint, obtained from one backward solve, turns inference over a prescribed source family into inner products (\S\ref{sec:lifted}). We impose three constraints that make its transpose usable, namely solver anchoring, closed-form quadratic centering and a block-triangular propagator, and show that they make the tangent linear model and the classical adjoint exact in the infinitesimal limit, leave a nonlinear remainder of second order in the amplitude, and give the learned blocks a well-posed, closed-form fit (\S\ref{sec:structure}). (ii) \textbf{One backward solve for every candidate source.} One backward solve per sensor yields a lifted adjoint whose physical-state part is exactly the classical adjoint. The linear term of the functional is therefore the classical positional embedding \citep{you2026positional}, and the learned channels add the finite-amplitude correction. Predicting the measurement of any candidate source, computing its gradient or sampling a full posterior then costs encoder passes instead of nonlinear solves. (iii) \textbf{Accurate, fast inference on three nonlinear PDE systems.} We test the framework on the viscous Burgers equation about an evolving base flow, a two-dimensional reaction--diffusion system, and two-dimensional Kolmogorov flow about a laminar shear, each with a prescribed low-dimensional source family. Against the solver, the classical adjoint and its quadratic correction, unconstrained lifts, structural ablations, a neural surrogate and networks trained on measurements, the lift is exact at small amplitude in both the measurement and its gradient, infers sources nearly as accurately as the solver, reproduces the solver's joint posterior on reaction--diffusion at a twentieth of the cost, and serves new sensor layouts without retraining (\S\ref{sec:experiments}).

\section{Related work}
Three lines of research are relevant to inferring finite-amplitude sources from sparse sensors. \textbf{Koopman and transfer operators.} Koopman theory represents nonlinear dynamics by a linear operator acting on observables of the state \citep{koopman1931hamiltonian, mezic2005spectral}, and its data-driven approximations are reviewed by \citet{brunton2022modern}. Extended dynamic mode decomposition fits a finite-dimensional Koopman matrix on a prescribed dictionary \citep{williams2015data}, and Koopman autoencoders learn the dictionary together with the matrix \citep{lusch2018deep, otto2019linearly}. The transpose of the EDMD matrix acts on the coefficients of observables, and together with the dictionary's Gram matrix it gives the Galerkin approximation of the Perron--Frobenius operator \citep{klus2016numerical}. Physics-informed DMD constrains the fitted operator to known structure such as symmetries and conservation laws \citep{baddoo2023physics}, and hybrid models add a learned correction to a known physical model \citep{yin2021augmenting, um2020solver}. Carleman linearization builds a linear embedding of a polynomial system from its equations, at the cost of an infinite hierarchy that must be truncated \citep{carleman1932application, kowalski1991nonlinear}. \textbf{Adjoints and learned surrogates.} Adjoint sensitivity is classical in fluid mechanics and control, where one backward solve gives the gradient of a measured functional with respect to every input \citep{errico1997adjoint, giles2000introduction, Luchini2014adjoint}. In turbulent flows the adjoint itself degrades over long horizons, which has motivated regularized and latent-space assimilation \citep{ke2026preconditioned, wang2026mitigating, gutierrez2026mixed, cleary2025latent}. Differentiable surrogates provide gradients through learned models \citep{holl2020learning, kochkov2021machine}, although these gradients are rarely validated against the solver. Sobolev training supervises the derivatives of a network directly, at the cost of tangent solves during training \citep{czarnecki2017sobolev}, and derivative-informed neural operators train on derivatives for inverse problems \citep{oleary2024derivative}. \textbf{Source inference.} Adjoint-based methods localize sources from sparse sensors with a model that is linear in the source amplitude, and a quadratic correction extends this linear embedding to second order \citep{wang2026localization}. Amortized simulation-based inference instead learns the posterior from simulations \citep{cranmer2020frontier}, and other reconstruction methods exploit the domain of dependence of each measurement \citep{you2026scalar, wang2025domain, gutierrez2026reconstructing, du2026inverse, dominguez2025inference}.

\section{Lifted sensitivity}
\label{sec:lifted}

This section poses the inference problem and shows how a Koopman lift turns its nonlinear measurement functional into an inner product with a lifted adjoint. Let $\state(t)\in\mathbb R^N$ be the discretized perturbation of the state about a base solution. A nonlinear solver advances it over a snapshot interval $\Delta s$, and we write this one-step map as $\state_{k+1}=\mathcal N_k(\state_k)$, $k=0,\dots,K-1$, with $T=K\Delta s$. We call an initial perturbation a source and write it as $\state_0=S(\boldsymbol p)$, where $\boldsymbol p$ collects a few parameters such as position, intensity, width, orientation and injection time. Because the dynamics are written for the perturbation about a base solution, the system is homogeneous, so a zero source produces a zero response at all times. We assume throughout that the base solution is known and that the solver can reproduce it. The measurements are linear functionals of the final state, $m_j=\langle\boldsymbol e_j,\state_K\rangle$ for $j=1,\dots,N_m$, where $\boldsymbol e_j$ is a point or footprint kernel. The solver also provides the linearized one-step propagators $\Lin_k=\partial\mathcal N_k/\partial\state$ about the base trajectory, which together form the discrete tangent linear model. The \emph{linear adjoint} $\qL$ carries a kernel backward through $\Lin_k^{\!\top}$, and it is what a classical adjoint code returns \citep{wang2019discrete, wang_wang_zaki_2022, wang2019spatial}. It is a single vector, so it cannot express the measurement $m_j(\boldsymbol p)$, which is a nonlinear function of $\boldsymbol p$ across the source family. We form the lifted adjoint once, about the base state, because the perturbation itself is the unknown that the inverse problem seeks. Section~\ref{sec:experiments} tests the functional and its gradient away from the base.

Let $\feat_\theta:\mathbb R^N\to\mathbb R^{cN}$ be a learned map to $c$ channels, and let $\lift(\state)=[\state;\feat_\theta(\state)]\in\mathbb R^{d}$, with $d=N(1+c)$, be the lifted state. The lifted linear model is $\Zs_{k+1}=\prop_k\Zs_k$ with the block-triangular propagator
\begin{equation}
\prop_k=\begin{bmatrix}\Lin_k & \prop_{q\Phi}\\ \boldsymbol 0 & \prop_{\Phi\Phi}\end{bmatrix},
\label{eq:G}
\end{equation}
where $\Lin_k$ is the solver's propagator, and $\prop_{q\Phi}$ and $\prop_{\Phi\Phi}$ are learned, time-invariant blocks that couple the features to the state and advance the features. Fitting all of $\prop_k$ by least squares on snapshot pairs would amount to extended dynamic mode decomposition \citep{williams2015data}. Here, by contrast, two of its blocks are imposed rather than fitted (\S\ref{sec:structure}). The lifted model starts from the encoded source, $\Zs_0=\lift(\state_0)$, and predicts the measurement as a linear functional of the lifted state, $\hat m_j=\langle[\boldsymbol e_j;\boldsymbol 0],\Zs_K\rangle$. Carrying this functional backward in time gives the \emph{lifted adjoint}
\begin{equation}
\begin{aligned}
\Zd_j(\tau_k)&=\prop_{K-k}^{\!\top}\cdots\prop_{K-1}^{\!\top}\begin{bmatrix}\boldsymbol e_j\\ \boldsymbol 0\end{bmatrix},\\
\hat m_j&=\big\langle \begin{bmatrix}\boldsymbol e_j\\ \boldsymbol 0\end{bmatrix},\Zs_K\big\rangle=\big\langle\Zd_j(\tau_k),\Zs_{K-k}\big\rangle
\end{aligned}
\label{eq:duality}
\end{equation}

for every backward time $\tau_k=k\Delta s$. The lifted adjoint holds the coefficients of the observable whose value on the lifted state at time $T-\tau$ equals the final measurement, just as the classical adjoint field is the sensor kernel carried backward by the transposed tangent model. The identity \eqref{eq:duality} holds exactly for the lifted rollout, so the only error of the lifted model relative to the solver is the error of that rollout, which Section~\ref{sec:experiments} measures. Three properties follow. First, $\Zd_j$ does not depend on the state, so it is computed once per sensor and reused for every source. Second, setting $\tau=T$ and $\Zs_0=\lift(S(\boldsymbol p))$ in \eqref{eq:duality} gives the predicted measurement of a finite source directly,
\begin{equation}
\hat m_j(\boldsymbol p)=\big\langle\Zd_{j,q}(T),S(\boldsymbol p)\big\rangle+\big\langle\Zd_{j,\Phi}(T),\feat_\theta(S(\boldsymbol p))\big\rangle,
\label{eq:family}
\end{equation}
a nonlinear functional of the source, which we call the \emph{nonlinear duality relation}. Its first term is the linear positional embedding of the classical adjoint, and its second term is the finite-amplitude correction. Evaluating it for a candidate costs one encoder pass and two inner products, with no time stepping. Third, by (C2) and (C3) the state block satisfies $\Zd_{j,q}(\tau_k)=\Lin_{K-k}^{\!\top}\cdots\Lin_{K-1}^{\!\top}\boldsymbol e_j=\qL$, which is the linear adjoint itself. Every inference in this paper scores candidates by evaluating the duality relation \eqref{eq:family} directly. Only the two-source point estimates of Appendix~\ref{app:twosource} are afterwards refined, by gradient descent on the same functional.

\textbf{Posterior density of a source.} The duality relation \eqref{eq:family} scores one candidate at a time, but the experiments also ask how precisely a source can be located, which a point estimate cannot tell us. We write $p=(x_s,y_s,I_s)$ for the parameters of one source and $\widehat{\boldsymbol m}(p_1,\dots,p_{n_s})$ for the measurement that a model predicts when $n_s$ sources are present together. Every model, including the solver, is evaluated on the combined field $\sum_aS(p_a)$, so that the prediction includes the interaction between the sources. With independent Gaussian noise of width $\sigma$ and a flat prior,
\begin{equation}
\begin{aligned}
\mathbb{P}(p_1,\dots,p_{n_s})&\;\propto\;
\exp\!\Big(-\tfrac{1}{2\sigma^{2}}\big\|\boldsymbol m_{\mathrm{obs}}
-\widehat{\boldsymbol m}(p_1,\dots,p_{n_s})\big\|^{2}\Big),
\\
\rho(x_s,y_s)&\;\propto\;
\int\!\mathrm{d}I_s\int\!\prod_{b\ge2}\mathrm{d}p_b\,
\mathbb{P}\big((x_s,y_s,I_s),p_2,\dots,p_{n_s}\big),
\end{aligned}
\label{eq:posterior}
\end{equation}
where $\rho$ is the density shown in the figures. Because the sources are exchangeable, $\rho$ has one peak per resolved source, and we estimate it from Markov chain Monte Carlo samples (Appendix~\ref{app:setups}).

The width $\sigma$ accounts for more than the observation noise. A surrogate's own error in predicting the measurement adds to the noise in the residual, and when this error dominates, a likelihood based on the noise alone is overconfident and can be confidently wrong. We therefore set $\sigma^{2}=\sigma_n^{2}+\sigma_{\mathrm{mod}}^{2}$, where $\sigma_{\mathrm{mod}}$ is the model's root-mean-square discrepancy from the solver on an independent set of combined pair fields. This makes models of different accuracy comparable, and for the solver $\sigma_{\mathrm{mod}}=0$.

\section{Trustworthy structure}
\label{sec:structure}
The lifted adjoint of Section~\ref{sec:lifted} is only as reliable as the transpose of the learned propagator. A lift fitted only for forward accuracy leaves its transpose unconstrained away from the states visited in training, and it overwrites the linear dynamics that the solver already knows, so it need not reduce to the classical adjoint at small amplitude. On Burgers about a quiescent base, for example, an unconstrained lift of the same size matches our forward accuracy, yet its rollout error stays between $4\%$ and $53\%$ as the amplitude vanishes instead of vanishing with it (Appendix~\ref{app:quiescent}). The three constraints below remove this failure by construction rather than through training.

\textbf{(C1) Quadratic centering.} The learned channels must be at least second order in the source perturbation, or they would corrupt the linear response that the solver already gives exactly. A nonzero $\feat_\theta(\boldsymbol 0)$ injects a constant through the coupling block, and a nonzero Jacobian at the base state adds a spurious linear term to the adjoint that does not vanish with the amplitude. We therefore require $\feat_\theta(\boldsymbol 0)=\boldsymbol 0$ and $\partial\feat_\theta/\partial\state|_{\boldsymbol 0}=\boldsymbol 0$, so that $\feat_\theta(\state)=O(\|\state\|^2)$. Given any smooth network $\rawfeat_\theta$, which is the encoder before centering, we set
\begin{equation}
\feat_\theta(\state)=\rawfeat_\theta(\state)-\rawfeat_\theta(\boldsymbol 0)-\boldsymbol J_0\state,\qquad \boldsymbol J_0=\partial \rawfeat_\theta/\partial\state|_{\boldsymbol 0},
\label{eq:centering}
\end{equation}
which is exact and differentiable in $\theta$. Both correction terms have closed forms for a tanh network (Appendix~\ref{app:setups}). For a translation-equivariant convolutional encoder, $\rawfeat_\theta(\boldsymbol 0)$ is one constant per channel and $\boldsymbol J_0$ is a fixed convolution, so the centering needs no extra pass of the encoder.

\textbf{(C2) Solver anchoring.} The state row of \eqref{eq:G} uses the exact linearized solver $\Lin_k$. 

\textbf{(C3) Block triangularity.} The feature rows depend on the features alone. The lower-left block is $\mathbf{0}$.

\paragraph{Identities.} Under (C1)--(C3), four identities hold for every $k$. (i) The $k$-step state block of the lifted propagator is exactly the solver's, $(\prop_{k-1}\cdots\prop_0)_{qq}=\Lin_{k-1}\cdots\Lin_0$, so at every horizon the linearization of the lifted model at the base state is the tangent linear model. (ii) The lifted rollout from $\state_0=\epsilon\boldsymbol v$ equals the tangent linear rollout up to $O(\epsilon^2)$, and the learned channels contribute only the $O(\epsilon^2)$ remainder. (iii) The state block of the lifted adjoint is the linear adjoint, $\Zd_{j,q}=\qL$, so the lifted functional reduces to the classical positional embedding as $\epsilon\to0$, since $\partial\feat_\theta/\partial\state=O(\epsilon)$. (iv) The fit of the learned blocks is a constrained ridge least-squares problem with a closed-form solution. Each proof follows in one line from the block structure and from $\feat_\theta=O(\|\state\|^2)$, and we verify the identities numerically to machine precision relative to the supplied $\Lin_k$, which on Burgers and Kolmogorov flow are central-difference Jacobians of the solver. None of (i)--(iii) holds for an unconstrained lift, whose linearization at the base state is a fitted matrix that matches the tangent model only where the data constrain it.

\paragraph{Closed-form fit and bilevel training.} Given snapshot pairs $(\state_k,\state_{k+1})$ along the training trajectories, with $\Lin_k\state_k$ supplied by the solver, the learned blocks solve
\begin{equation}
\begin{aligned}
\prop_{q\Phi}&=\arg\min\textstyle\sum_k\|\state_{k+1}-\Lin_k\state_k-\prop_{q\Phi}\feat_\theta(\state_k)\|^2+\lambda\|\cdot\|^2,\\
\prop_{\Phi\Phi}&=\arg\min\textstyle\sum_k\|\feat_\theta(\state_{k+1})-\prop_{\Phi\Phi}\feat_\theta(\state_k)\|^2+\lambda\|\cdot\|^2,
\end{aligned}
\label{eq:fit}
\end{equation}
Each is a single Gram solve on the feature block, with rows scaled to unit RMS and a ridge parameter $\lambda$ set relative to the trace. The data only have to explain the residual $\state_{k+1}-\Lin_k\state_k$, because the linear transport is not learned. We train the encoder parameters $\theta$ by gradient descent on the multi-step rollout error of the lifted model, differentiating through the closed-form solve of the inner loop, and we lengthen the rollout gradually during training. Every problem and every ablation uses the same trainer.

\paragraph{Baselines with the same code.} Removing (C1) leaves raw, uncentered features (anchoring only). Removing (C2) and (C3) fits the full $\prop$ by ridge EDMD (centering only). Removing all three gives the unconstrained Koopman autoencoder. A neural surrogate uses the same encoder with a linear head as a residual one-step map, $\state_{k+1}=\Lin_k\state_k+\boldsymbol W \rawfeat_\theta(\state_k)+\boldsymbol b$, trained on the same rollout loss. Its gradient is exact by automatic differentiation, but scoring each candidate requires a rollout.

\section{Experiments}
\label{sec:experiments}
We evaluate \textsc{SALTO} on three systems, each showing a different way in which a nonlinearity defeats a linear adjoint. The first has self-advection that carries the response away from where the linear model places it, the second has a saturating reaction, and the third has a vortex that advects itself inside a sheared Navier--Stokes flow. Each system has a prescribed low-dimensional source family, and we compare against the solver and two classical models. The linear embedding predicts a measurement as $\langle\qL,S(\boldsymbol p)\rangle$, and the quadratic correction adds the second-order term of the expansion about the base, computed with second-order tangent and adjoint solves \citep{wang2026localization}. Throughout, $\varepsilon_m$ is the relative error of a model's predicted measurements at the top of the training range. The median position error $\delta_x$ and the median relative intensity error $\delta_I$ describe the one-shot inversion at $1\%$ noise, with $\delta_x$ in units of the candidate spacing so that it compares across domains. The cost is the time for one estimate, which is a single pass over the candidate grid. Because candidate measurements do not depend on the data, a dictionary of solver measurements could also be cached once per sensor layout and horizon. The lift's cost advantage therefore lies in what no finite dictionary covers, such as the continuous joint posterior of several sources. Table~\ref{tab:main} collects accuracy and cost, with bootstrap intervals in Appendix~\ref{app:tables}, for one lift per system, with $c=24$ channels on Burgers and $c=4$ on the two-dimensional systems. The paper uses three kinds of lift, namely the $c=4$ dense lifts of Table~\ref{tab:main}, Fig.~\ref{fig:kolmores} and the reuse study, the $c=2$ dense lifts of the ablations, evaluated on a separate draw that extends down to zero intensity, and an eight-channel translation-equivariant lift trained on random fields for Fig.~\ref{fig:rdres}. Appendix~\ref{app:setups} gives the equations, discretizations, source families, sensors, architectures and hardware.

\begin{table}[t]
\centering\small
\begin{tabular}{@{}ll ccc c@{}}\toprule
 & & \multicolumn{3}{c}{accuracy} & cost \\
\cmidrule(lr){3-5}\cmidrule(lr){6-6}
case & model & $\varepsilon_m$ (\%) & $\delta_x$ (cells) & $\delta_I$ (\%) & seconds \\ \midrule
\multirow{4}{*}{\begin{tabular}[c]{@{}l@{}}Burgers\\ \emph{self-advection}\end{tabular}} & linear & 72.9 & 6.00 & 15.2 & 0.003 \\
 & quadratic & 96.1 & 3.00 & 8.9 & 4.1 \\
 & lifted (ours) & \textbf{1.8} & \textbf{1.00} & \textbf{3.6} & 9.6 \\
 & solver & -- & 1.00 & 3.4 & 167 \\
\midrule
\multirow{4}{*}{\begin{tabular}[c]{@{}l@{}}reaction--diffusion\\ \emph{saturation}\end{tabular}} & linear & 164.5 & 0.63 & 40.2 & 0.005 \\
 & quadratic & 174.5 & 0.91 & 37.1 & 1.9 \\
 & lifted (ours) & \textbf{2.4} & \textbf{0.55} & \textbf{14.2} & 11.9 \\
 & solver & -- & 0.46 & 14.4 & 21.3 \\
\midrule
\multirow{4}{*}{\begin{tabular}[c]{@{}l@{}}Kolmogorov flow\\ \emph{sheared advection}\end{tabular}} & linear & 41.3 & 0.63 & 7.2 & 0.008 \\
 & quadratic & 22.6 & 0.45 & 6.3 & 31.0 \\
 & lifted (ours) & \textbf{7.7} & \textbf{0.41} & \textbf{3.9} & 11.8 \\
 & solver & -- & 0.41 & 2.8 & 214 \\
\bottomrule\end{tabular}

\caption{Accuracy and cost on the three systems, with the $c=24$ lift on Burgers and the $c=4$ lifts on the two-dimensional systems. Each block uses one common draw of test sources. Bold marks the best of the three models, the solver serves as the reference, and timings are wall-clock times.}
\label{tab:main}
\end{table}

\subsection{Self-advection: Burgers about an evolving base flow}
\label{sec:burgers}

\begin{figure}[ht]
\centering\includegraphics[width=\linewidth]{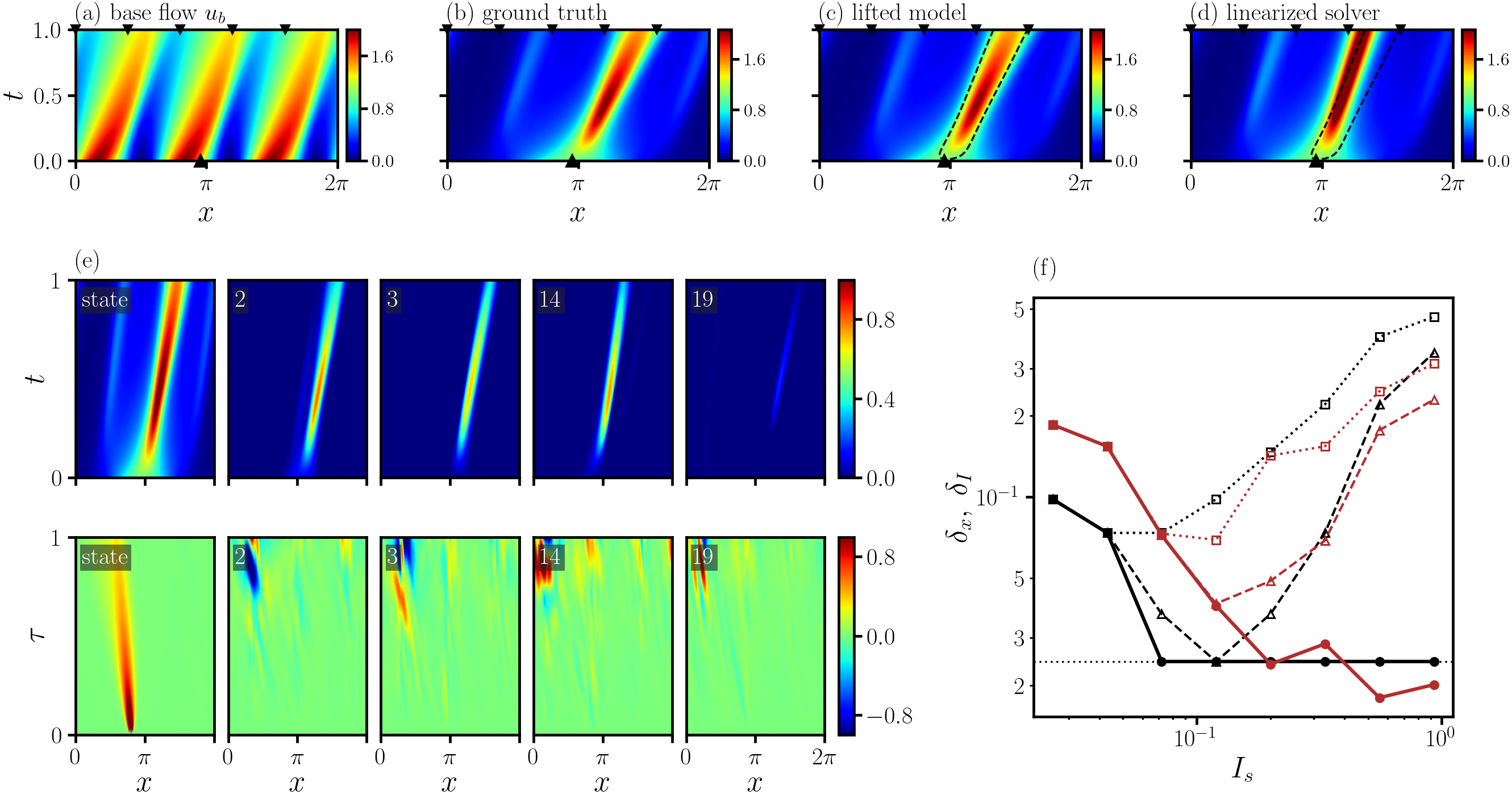}
\caption{\textbf{(a--d)} Space--time evolution of the base flow and of one release, from the solver, the lifted model and the linearized solver. The dashed line is the half-maximum contour of the ground truth. \textbf{(e)} Five representative blocks of the lift, forward in time (top) and adjoint in backward time $\tau$ (bottom). \textbf{(f)} Median inversion error for 800 sources at $1\%$ noise, with $\delta_x$ in black and $\delta_I$ in red, and linear dotted, quadratic dashed and lifted solid. The dotted horizontal line marks one grid cell.}
\label{fig:burgersres}
\end{figure}
The first system is the viscous Burgers equation perturbed about an evolving base flow (Appendix~\ref{app:setups}). The linearized solver misses both the steepening and the slowing of the front (Fig.~\ref{fig:burgersres}). Its error at $T$ grows from about a tenth of the response for the weakest source to four fifths for the strongest, whereas the lifted model keeps it below a few per cent. When we invert $(x_s,I_s)$ in one shot from the five sensors by a grid search with $1\%$ noise, the lifted functional matches brute-force inversion through the solver in intensity in every band and is at least as accurate in position (Appendix~\ref{app:tables}). The quadratic correction is the stronger classical baseline for inference. It halves the position error of the linear embedding and reduces its intensity error from $15\%$ to $9\%$, although its measurement error at the top of the training range is larger than the linear embedding's. It still falls short of the lifted functional, because it expands about the base, and the front carries the release beyond the amplitude at which that expansion is accurate.

\subsection{Saturation: two-dimensional reaction--diffusion}
\label{sec:rd}\label{sec:inference}
The second system saturates rather than advects, and we use it to study inference in full, from a single source to the joint posterior of two (Appendix~\ref{app:setups}). \textbf{Forward, measurement, small amplitude.} The four-channel lifted functional predicts the nine measurements to within $2.4\%$ at the top of the training range and $4\%$ beyond it, where the error of the linear embedding exceeds the measurement itself. As the amplitude decreases, the error of the two-channel ablation lift decreases in proportion, reaching $0.008\%$ at $I_s=10^{-3}$, below the linear embedding's $0.06\%$. There the unconstrained lift is off by $23\%$, and the neural surrogate by more than the measurement itself (Appendix~\ref{app:tables}).

\textbf{Inference and cost.} We infer each source by a grid search over $13{,}824$ candidates, scoring each with the duality relation \eqref{eq:family}, with $1\%$ noise added to the measurements. Over 60 test sources the lifted functional recovers the position to $0.55$ cells at the median, close to the $0.46$ cells of brute-force inversion through the solver, and matches its intensity error of $14\%$. It places 48 of the 60 sources within one cell of the truth, against 49 for brute force and 39 for the linear embedding (Appendix~\ref{app:tables}). Both classical baselines fall behind. The linear embedding cannot represent saturation and underestimates the intensity by $40\%$ at the median, reading a saturated response as a weaker source, and the quadratic correction helps only within its radius of convergence, since the expansion diverges for the stronger sources. On this problem, scoring by inner products is the most accurate of the surrogates but not the cheapest. It takes twelve seconds per estimate, against two for the quadratic correction and twenty-one for the solver (Table~\ref{tab:main}).

\textbf{Two sources at once.} When two sources overlap, one to three source widths apart, their joint measurement differs from the sum of their separate measurements by about a sixth (Appendix~\ref{app:twosource}). The lift used here is more general than the one reported in the tables. Its learned blocks are translation-equivariant (Appendix~\ref{app:equivariant}), it has eight channels, and it is trained on 1600 band-limited positive random fields rather than on the blob family that the inference searches over, so it never sees a source of the kind it is asked to find. The training fields must stay positive, because the reaction $-\kappa u-\gamma u^2$ drives $u$ without bound once $u<-\kappa/\gamma$.

\begin{figure}[t]
\centering\includegraphics[width=\linewidth]{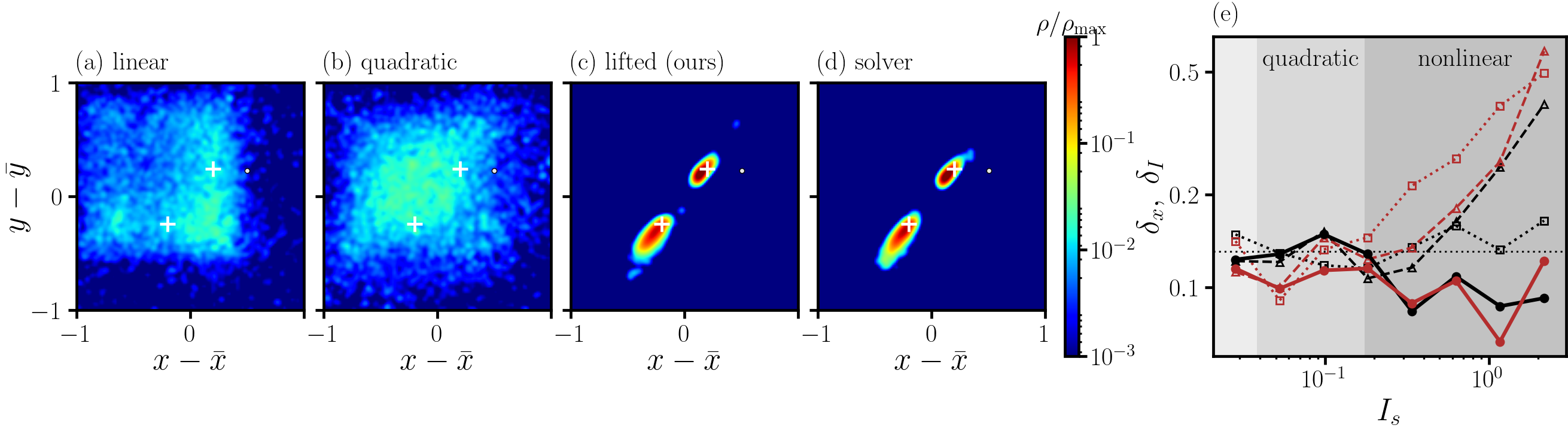}
\caption{Two-dimensional reaction--diffusion, with the eight-channel equivariant lift. The lifted posterior reproduces the solver's, while the classical ones spread over the whole window. \textbf{(a--d)} Joint posterior density of source position (Eq.~\ref{eq:posterior}) for a pair $1.4$ source widths apart, on a logarithmic scale. Each model is evaluated on the combined field, and its likelihood is widened by its own discrepancy. Crosses mark the true sources and the dot marks a probe. \textbf{(e)} Error of the one-shot inversion against the true intensity, with $\delta_x$ in black and $\delta_I$ in red, and linear dotted, quadratic dashed and lifted solid. Shading marks where each classical model still predicts the measurements to within three per cent, and the narrow unlabelled band at the left is where the linear embedding does.}
\label{fig:rdres}
\end{figure}

A point estimate cannot tell whether a pair has been resolved, so we sample the joint posterior of both sources (Eq.~\ref{eq:posterior}), applying every model to the combined field (Fig.~\ref{fig:rdres}a--d). Over 17 overlapping pairs the lifted posterior reproduces the solver's. Its mean lies a median of $0.4$ solver posterior standard deviations from the solver's mean. The central $95\%$ interval of each of the six parameters contains the true value in $87\%$ of the 102 cases, against $85\%$ for the solver under the same sampler, a difference within the binomial standard error of about $3.5$ points. Each lifted posterior costs $13$ seconds, against $295$ for the solver. The linear and quadratic posteriors are $13$ to $27$ times wider, with means $3.6$ and $5.6$ solver standard deviations away, so their full coverage only reflects that they are uninformative. Although this lift was trained on generic random fields and never on the blob family, it matches the solver on that family.

\subsection{Sheared advection: Kolmogorov flow}
\label{sec:kolmo}
\begin{figure}[t]
\centering\includegraphics[width=\linewidth]{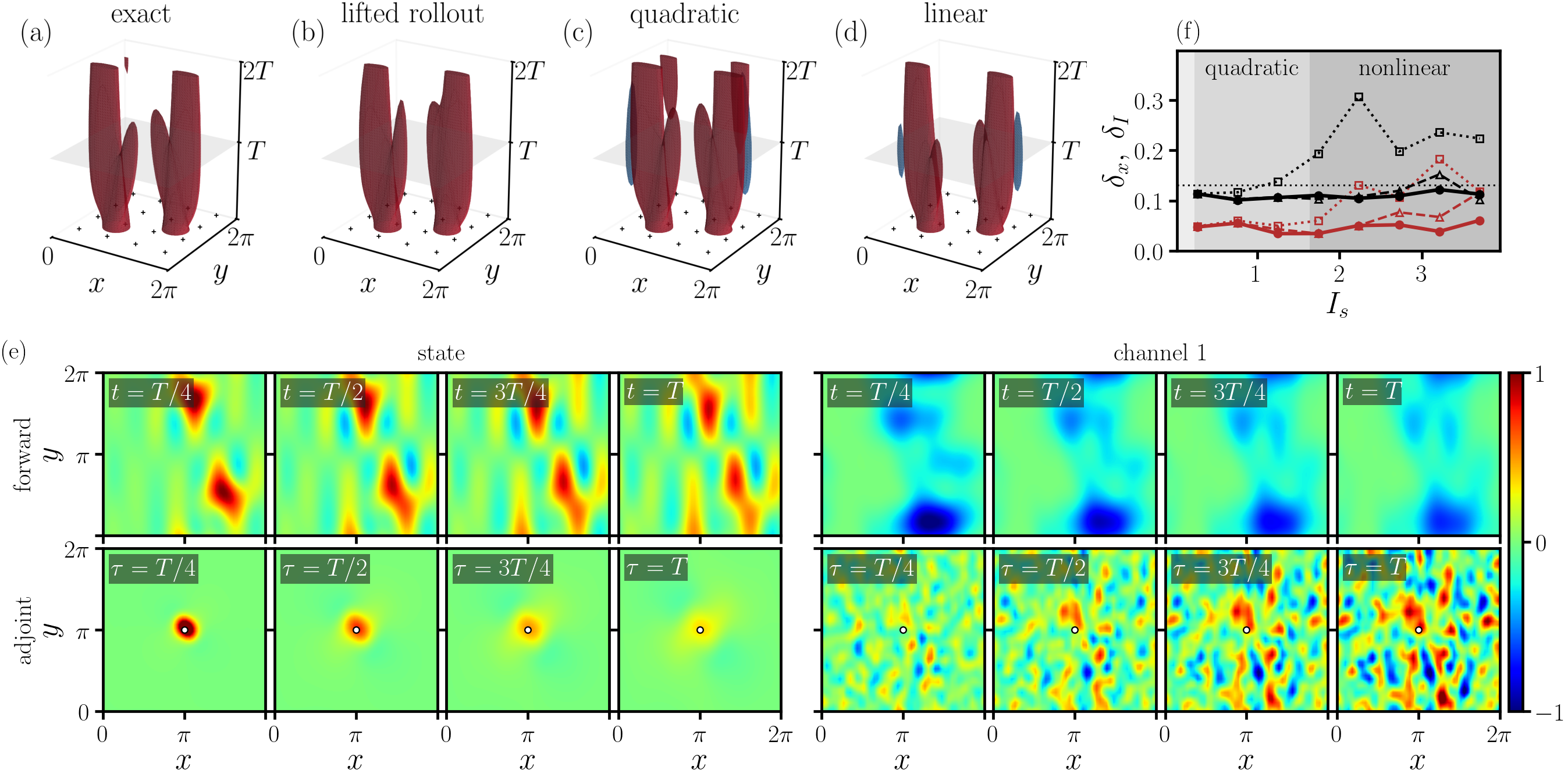}
\caption{Kolmogorov flow. \textbf{(a--d)} Perturbation vorticity at $\pm35\%$ of its initial maximum, time upward to $2T$, for the most strongly interacting pair, from the solver, the lifted rollout, the quadratic expansion and the linear superposition. \textbf{(e)} The state and one learned channel along that pair's solver trajectory (top), and the same blocks of the centre probe's lifted adjoint $\Zd$ (circle) in backward time (bottom), each on its own scale. The state block is exactly $\qL$. The channel block is spread out because a measurement uses only its projection on the channels. \textbf{(f)} Inversion error over 600 further sources, as in Fig.~\ref{fig:rdres}e on linear axes; the narrow band is where the linear embedding holds.}
\label{fig:kolmores}
\end{figure}
The third system is the two-dimensional Navier--Stokes equations in vorticity form about a steady laminar shear flow, with a localized injection of vorticity as the source (Appendix~\ref{app:setups}). The linearized propagator already carries the injected vorticity downstream and tilts it with the shear, so the lift only has to learn how the injected vortex advects itself (Fig.~\ref{fig:kolmores}e). At the top of the training range, missing this self-advection costs the linear embedding $41\%$ of the measurement and the quadratic correction $23\%$, against $7.7\%$ for the lifted functional (Table~\ref{tab:main}).

\textbf{Inference and cost.} The one-shot inversion of the 60 test sources matches the solver in position ($0.41$ cells) and nearly matches it in intensity ($3.9\%$ against $2.8\%$), whereas the linear embedding reaches $0.63$ cells and $7.2\%$ (Table~\ref{tab:main}, Fig.~\ref{fig:kolmores}f). Scoring the $13{,}824$ candidates takes $12$ seconds with the lifted functional, against $214$ for the solver and $31$ for the quadratic correction. For separated strong pairs the lift places both sources within one cell in $38\%$ of cases, against $13\%$ for the linear embedding and all pairs for brute force (Appendix~\ref{app:twosource}).

\begin{table}[t]
\centering\small
\caption{What the constraints buy (two-channel lifts). Only the anchored, centered lift is exact as the amplitude vanishes, in the measurement at $I_s=10^{-3}$ (a ratio) and in the gradient for $I_s\in[0.01,0.1)$, and on Kolmogorov flow it is the most accurate learned model throughout. The neural surrogate did not reach the lift's training loss on reaction--diffusion. Bold marks the best per column.}
\label{tab:ablation}
\begingroup\setlength{\tabcolsep}{3.5pt}\begin{tabular}{@{}l cccc cccc@{}}\toprule
 & \multicolumn{4}{c}{reaction--diffusion} & \multicolumn{4}{c}{Kolmogorov flow} \\
\cmidrule(lr){2-5}\cmidrule(lr){6-9}
 & $\varepsilon_m$ (\%) & $\varepsilon_m$ & \multicolumn{2}{c}{gradient (\%)} & $\varepsilon_m$ (\%) & $\varepsilon_m$ & \multicolumn{2}{c}{gradient (\%)} \\
\cmidrule(lr){4-5}\cmidrule(lr){8-9}
model & top band & $I_s{=}10^{-3}$ & weak & train & top band & $I_s{=}10^{-3}$ & weak & train \\ \midrule
lifted (ours) & 6.0 & \textbf{$8\times10^{-5}$} & \textbf{0.7} & 21 & \textbf{13} & \textbf{$9\times10^{-5}$} & \textbf{0.6} & \textbf{17} \\
unconstrained & \textbf{3.2} & 0.23 & 4.9 & \textbf{7.8} & 14 & 0.67 & 16 & 25 \\
centering only & 4.8 & $3\times10^{-2}$ & 2.9 & 17 & 14 & $6\times10^{-2}$ & 10 & 20 \\
anchoring only & 9.2 & 1.9 & 11 & 23 & 22 & 8.2 & 17 & 27 \\
neural surrogate & 51 & 124 & 66 & 94 & 34 & 9 & 18 & 37 \\
linear embedding & 147 & $6\times10^{-4}$ & 5.1 & 205 & 40 & $2\times10^{-4}$ & 1.1 & 43 \\
\bottomrule\end{tabular}\endgroup

\end{table}
\textbf{Ablations.} The constraints make the lift exact at small amplitude, where every unconstrained variant fails, while keeping its finite-amplitude accuracy competitive (Table~\ref{tab:ablation}). At $I_s=10^{-3}$ on reaction--diffusion the unconstrained lift is off by $23\%$ against $0.008\%$ in the current study, because its fitted linear part does not reduce to the solver's. On Kolmogorov flow ours is the most accurate learned model in every column. Its measurement gradient is also within $0.7\%$ of the solver's for the weakest sources, which no other learned model achieves.

\textbf{What the lifted dynamics add.} The lift learns its features through the dynamics, and these features carry the finite-amplitude response. On Burgers about a quiescent base, a Carleman lift built on twelve POD modes is five times less accurate than the learned lift (Appendix~\ref{app:quiescent}). Trained on the same simulations, a network of the same architecture fitted to the measurements alone is less accurate in every band on both systems, by up to a factor of twenty, and a network from the source parameters is off by $13$--$130\%$ for weak sources and cannot represent two sources (Appendix~\ref{app:tables}). Because the propagator is shared, one lift serves any sensor layout and time with a single backward solve, even beyond its training horizon, where at $1.5T$ it stays within $13\%$ on reaction--diffusion and $17\%$ on Kolmogorov flow against up to $196\%$ and $63\%$ for the linear embedding (Appendix~\ref{app:tables}).

\section{Discussion and outlook}
We have proposed \textsc{SALTO}, which performs linear sensitivity analysis in the lifted Koopman space to obtain a reusable nonlinear sensitivity in the original space, exact at small amplitude by construction, and which opens the way to efficient and accurate causal inference. Open questions include how many lifted channels and which architecture a problem requires, how the lift converges as it grows, and how far the idea extends to chaotic turbulent flows, which are closer to real applications.

\section*{AI Use Statement}

Generative AI tools were used to assist with editing and polishing the
manuscript, including improving clarity, grammar, presentation, and refining the visualization and drawing part of the schematics. All
AI-assisted text was reviewed and verified by the authors, who take full
responsibility for the final content of the paper.

\bibliography{refs}

\begin{thebibliography}{35}
\providecommand{\natexlab}[1]{#1}
\providecommand{\url}[1]{\texttt{#1}}
\expandafter\ifx\csname urlstyle\endcsname\relax
  \providecommand{\doi}[1]{doi: #1}\else
  \providecommand{\doi}{doi: \begingroup \urlstyle{rm}\Url}\fi

\bibitem[Baddoo et~al.(2023)Baddoo, Herrmann, McKeon, Kutz, and
  Brunton]{baddoo2023physics}
Peter~J. Baddoo, Benjamin Herrmann, Beverley~J. McKeon, J.~Nathan Kutz, and
  Steven~L. Brunton.
\newblock Physics-informed dynamic mode decomposition.
\newblock \emph{Proceedings of the Royal Society A}, 479\penalty0
  (2271):\penalty0 20220576, 2023.

\bibitem[Brunton et~al.(2022)Brunton, Budi{\v{s}}i{\'c}, Kaiser, and
  Kutz]{brunton2022modern}
Steven~L Brunton, Marko Budi{\v{s}}i{\'c}, Eurika Kaiser, and J~Nathan Kutz.
\newblock Modern {K}oopman theory for dynamical systems.
\newblock \emph{SIAM Review}, 64\penalty0 (2):\penalty0 229--340, 2022.

\bibitem[Carleman(1932)]{carleman1932application}
Torsten Carleman.
\newblock Application de la th{\'e}orie des {\'e}quations int{\'e}grales
  lin{\'e}aires aux syst{\`e}mes d'{\'e}quations diff{\'e}rentielles non
  lin{\'e}aires.
\newblock \emph{Acta Mathematica}, 59:\penalty0 63--87, 1932.

\bibitem[Cleary et~al.(2025)Cleary, Wang, and Zaki]{cleary2025latent}
Andrew Cleary, Qi~Wang, and Tamer~A Zaki.
\newblock Latent-space variational data assimilation in two-dimensional
  turbulence.
\newblock \emph{arXiv preprint arXiv:2512.15470}, 2025.

\bibitem[Cranmer et~al.(2020)Cranmer, Brehmer, and Louppe]{cranmer2020frontier}
Kyle Cranmer, Johann Brehmer, and Gilles Louppe.
\newblock The frontier of simulation-based inference.
\newblock \emph{Proceedings of the National Academy of Sciences}, 117\penalty0
  (48):\penalty0 30055--30062, 2020.

\bibitem[Czarnecki et~al.(2017)Czarnecki, Osindero, Jaderberg, Swirszcz, and
  Pascanu]{czarnecki2017sobolev}
Wojciech~M Czarnecki, Simon Osindero, Max Jaderberg, Grzegorz Swirszcz, and
  Razvan Pascanu.
\newblock Sobolev training for neural networks.
\newblock In \emph{Advances in Neural Information Processing Systems},
  volume~30, 2017.

\bibitem[Dom{\'\i}nguez-V{\'a}zquez et~al.(2025)Dom{\'\i}nguez-V{\'a}zquez,
  Escobar-Castaneda, Wang, and Jacobs]{dominguez2025inference}
Daniel Dom{\'\i}nguez-V{\'a}zquez, Nicolas Escobar-Castaneda, Qi~Wang, and
  Gustaaf~B Jacobs.
\newblock Inference of inertial particle dynamics from limited measurements.
\newblock \emph{Physics of Fluids}, 37\penalty0 (4), 2025.

\bibitem[Du \& Wang(2026)Du and Wang]{du2026inverse}
Yifan Du and Qi~Wang.
\newblock Inverse reconstruction of free-stream vortices using surface pressure
  on a joukovsky airfoil.
\newblock In \emph{AIAA SCITECH 2026 Forum}, pp.\  1112, 2026.

\bibitem[Errico(1997)]{errico1997adjoint}
Ronald~M. Errico.
\newblock What is an adjoint model?
\newblock \emph{Bulletin of the American Meteorological Society}, 78\penalty0
  (11):\penalty0 2577--2591, 1997.

\bibitem[Giles \& Pierce(2000)Giles and Pierce]{giles2000introduction}
Michael~B. Giles and Niles~A. Pierce.
\newblock An introduction to the adjoint approach to design.
\newblock \emph{Flow, Turbulence and Combustion}, 65\penalty0 (3):\penalty0
  393--415, 2000.

\bibitem[Gutierrez \& Wang(2026{\natexlab{a}})Gutierrez and
  Wang]{gutierrez2026mixed}
Leopoldo Gutierrez and Qi~Wang.
\newblock Mixed lagrangian-eulerian data assimilation for turbulent channel
  flows.
\newblock In \emph{AIAA SciTech 2026 Forum}, pp.\  0682, 2026{\natexlab{a}}.

\bibitem[Gutierrez \& Wang(2026{\natexlab{b}})Gutierrez and
  Wang]{gutierrez2026reconstructing}
Leopoldo Gutierrez and Qi~Wang.
\newblock Reconstructing dangerous flow events using the domain of dependence
  from surface pressure data.
\newblock In \emph{AIAA SCITECH 2026 Forum}, pp.\  0510, 2026{\natexlab{b}}.

\bibitem[Holl et~al.(2020)Holl, Koltun, and Thuerey]{holl2020learning}
Philipp Holl, Vladlen Koltun, and Nils Thuerey.
\newblock Learning to control {PDE}s with differentiable physics.
\newblock In \emph{International Conference on Learning Representations}, 2020.

\bibitem[Ke et~al.(2026)Ke, You, and Wang]{ke2026preconditioned}
Hongyi Ke, Zejian You, and Qi~Wang.
\newblock Preconditioned adjoint data assimilation for two-dimensional decaying
  and forced turbulence.
\newblock \emph{arXiv preprint arXiv:2602.14016}, 2026.

\bibitem[Klus et~al.(2016)Klus, Koltai, and Sch{\"u}tte]{klus2016numerical}
Stefan Klus, P{\'e}ter Koltai, and Christof Sch{\"u}tte.
\newblock On the numerical approximation of the {P}erron--{F}robenius and
  {K}oopman operator.
\newblock \emph{Journal of Computational Dynamics}, 3\penalty0 (1):\penalty0
  51--79, 2016.

\bibitem[Kochkov et~al.(2021)Kochkov, Smith, Alieva, Wang, Brenner, and
  Hoyer]{kochkov2021machine}
Dmitrii Kochkov, Jamie~A. Smith, Ayya Alieva, Qing Wang, Michael~P. Brenner,
  and Stephan Hoyer.
\newblock Machine learning-accelerated computational fluid dynamics.
\newblock \emph{Proceedings of the National Academy of Sciences}, 118\penalty0
  (21):\penalty0 e2101784118, 2021.

\bibitem[Koopman(1931)]{koopman1931hamiltonian}
Bernard~O Koopman.
\newblock Hamiltonian systems and transformation in hilbert space.
\newblock \emph{Proceedings of the National Academy of Sciences}, 17\penalty0
  (5):\penalty0 315--318, 1931.

\bibitem[Kowalski \& Steeb(1991)Kowalski and Steeb]{kowalski1991nonlinear}
Krzysztof Kowalski and Willi-Hans Steeb.
\newblock \emph{Nonlinear Dynamical Systems and {C}arleman Linearization}.
\newblock World Scientific, 1991.

\bibitem[Luchini \& Bottaro(2014)Luchini and Bottaro]{Luchini2014adjoint}
Paolo Luchini and Alessandro Bottaro.
\newblock Adjoint equations in stability analysis.
\newblock \emph{Annual Review of fluid mechanics}, 46:\penalty0 493--517, 2014.

\bibitem[Lusch et~al.(2018)Lusch, Kutz, and Brunton]{lusch2018deep}
Bethany Lusch, J~Nathan Kutz, and Steven~L Brunton.
\newblock Deep learning for universal linear embeddings of nonlinear dynamics.
\newblock \emph{Nature Communications}, 9\penalty0 (1):\penalty0 4950, 2018.

\bibitem[Mezi{\'c}(2005)]{mezic2005spectral}
Igor Mezi{\'c}.
\newblock Spectral properties of dynamical systems, model reduction and
  decompositions.
\newblock \emph{Nonlinear Dynamics}, 41\penalty0 (1):\penalty0 309--325, 2005.

\bibitem[O'Leary-Roseberry et~al.(2024)O'Leary-Roseberry, Chen, Villa, and
  Ghattas]{oleary2024derivative}
Thomas O'Leary-Roseberry, Peng Chen, Umberto Villa, and Omar Ghattas.
\newblock Derivative-informed neural operator: an efficient framework for
  high-dimensional parametric derivative learning.
\newblock \emph{Journal of Computational Physics}, 496:\penalty0 112555, 2024.

\bibitem[Otto \& Rowley(2019)Otto and Rowley]{otto2019linearly}
Samuel~E Otto and Clarence~W Rowley.
\newblock Linearly recurrent autoencoder networks for learning dynamics.
\newblock \emph{SIAM Journal on Applied Dynamical Systems}, 18\penalty0
  (1):\penalty0 558--593, 2019.

\bibitem[Um et~al.(2020)Um, Brand, Fei, Holl, and Thuerey]{um2020solver}
Kiwon Um, Robert Brand, Yun~Raymond Fei, Philipp Holl, and Nils Thuerey.
\newblock Solver-in-the-loop: Learning from differentiable physics to interact
  with iterative {PDE}-solvers.
\newblock In \emph{Advances in Neural Information Processing Systems},
  volume~33, pp.\  6111--6122, 2020.

\bibitem[Wang et~al.(2019{\natexlab{a}})Wang, Wang, and Zaki]{wang2019discrete}
Mengze Wang, Qi~Wang, and Tamer~A Zaki.
\newblock Discrete adjoint of fractional-step incompressible navier-stokes
  solver in curvilinear coordinates and application to data assimilation.
\newblock \emph{Journal of Computational Physics}, 396:\penalty0 427--450,
  2019{\natexlab{a}}.

\bibitem[Wang \& You(2026)Wang and You]{wang2026localization}
Qi~Wang and Zejian You.
\newblock Localization of sources in weakly nonlinear fluid systems using
  linear and quadratic sensitivity analysis.
\newblock \emph{Physical Review Fluids}, 11\penalty0 (5):\penalty0 054901,
  2026.

\bibitem[Wang \& Zaki(2025)Wang and Zaki]{wang2025domain}
Qi~Wang and Tamer~A Zaki.
\newblock Domain of dependence for wall-pressure measurements in high-speed
  boundary layers.
\newblock \emph{Journal of Fluid Mechanics}, 1009:\penalty0 A67, 2025.

\bibitem[Wang \& Zaki(2026)Wang and Zaki]{wang2026mitigating}
Qi~Wang and Tamer~A Zaki.
\newblock Mitigating adjoint chaos in wall turbulence.
\newblock \emph{Journal of Fluid Mechanics}, 1039:\penalty0 A48, 2026.

\bibitem[Wang et~al.(2019{\natexlab{b}})Wang, Hasegawa, and
  Zaki]{wang2019spatial}
Qi~Wang, Yosuke Hasegawa, and Tamer~A Zaki.
\newblock Spatial reconstruction of steady scalar sources from remote
  measurements in turbulent flow.
\newblock \emph{Journal of Fluid Mechanics}, 870:\penalty0 316--352,
  2019{\natexlab{b}}.

\bibitem[Wang et~al.(2022)Wang, Wang, and Zaki]{wang_wang_zaki_2022}
Qi~Wang, Mengze Wang, and Tamer~A. Zaki.
\newblock What is observable from wall data in turbulent channel flow?
\newblock \emph{Journal of Fluid Mechanics}, 941:\penalty0 A48, 2022.
\newblock \doi{10.1017/jfm.2022.295}.

\bibitem[Williams et~al.(2015)Williams, Kevrekidis, and
  Rowley]{williams2015data}
Matthew~O Williams, Ioannis~G Kevrekidis, and Clarence~W Rowley.
\newblock A data--driven approximation of the koopman operator: Extending
  dynamic mode decomposition.
\newblock \emph{Journal of Nonlinear Science}, 25:\penalty0 1307--1346, 2015.

\bibitem[Yin et~al.(2021)Yin, Le~Guen, Dona, de~B{\'e}zenac, Ayed, Thome, and
  Gallinari]{yin2021augmenting}
Yuan Yin, Vincent Le~Guen, J{\'e}r{\'e}mie Dona, Emmanuel de~B{\'e}zenac,
  Ibrahim Ayed, Nicolas Thome, and Patrick Gallinari.
\newblock Augmenting physical models with deep networks for complex dynamics
  forecasting.
\newblock In \emph{International Conference on Learning Representations}, 2021.

\bibitem[You \& Wang(2026)You and Wang]{you2026positional}
Zejian You and Qi~Wang.
\newblock Positional embeddings of scalar sources in turbulent channel flow
  from linear adjoint sensitivity.
\newblock \emph{in preparation}, 2026.

\bibitem[You et~al.(2026)You, Wang, and Zhu]{you2026scalar}
Zejian You, Qi~Wang, and Xiaowei Zhu.
\newblock Scalar source localization using multi-sensor domains of dependence
  in turbulent channel flow.
\newblock In \emph{AIAA SCITECH 2026 Forum}, pp.\  2343, 2026.

\bibitem[Zaki(2025)]{zaki2025arfm}
Tamer~A. Zaki.
\newblock Turbulence from an observer perspective.
\newblock \emph{Annual Review of Fluid Mechanics}, 57:\penalty0 311--334, 2025.
\newblock ISSN 1545-4479.
\newblock \doi{https://doi.org/10.1146/annurev-fluid-030424-114735}.
\newblock URL
  \url{https://www.annualreviews.org/content/journals/10.1146/annurev-fluid-030424-114735}.

\end{thebibliography}
\bibliographystyle{iclr2026_conference}

\appendix
\setcounter{topnumber}{3}\setcounter{totalnumber}{5}\renewcommand{\topfraction}{0.85}\renewcommand{\textfraction}{0.15}\renewcommand{\floatpagefraction}{0.8}
\section{Problem setups and protocol}
\label{app:setups}
This appendix gives everything needed to reproduce the results, namely the three systems, the inference protocol and training, the posterior sampler and the translation-equivariant blocks. We chose the systems for the kind of nonlinearity each one presents rather than for their physics, and they appear in the same order as in Section~\ref{sec:experiments}.

\subsection{The three systems}
\textbf{Self-advection: Burgers about an evolving base flow.} We perturb the viscous Burgers equation $u_t+uu_x=\nu u_{xx}$ on $[0,2\pi)$, with $\nu=0.1$, about the evolving base flow that starts from $u_b(x,0)=1+\sin3x$. This is the configuration of an earlier linear-adjoint localization study \citep{wang2026localization}. The grid has $N=256$ points, the horizon $T=1$ is divided into $K=50$ intervals, and the time-varying propagators $\Lin_k$ are obtained from the solver by central differences. The source is a Gaussian of unit width at position $x_s$ with intensity $I_s$, and five point sensors spread across the domain record it. The encoder is a periodic dilated convolutional network with $c=24$ channels, trained on 800 releases of one to three bumps with $I_s\le0.7$.

\textbf{Saturation: two-dimensional reaction--diffusion.} The dynamics are $u_t=D\nabla^2u-\kappa u-\gamma u^2+\boldsymbol c\cdot\nabla u$, with $D=0.05$, $\kappa=0.2$, $\gamma=1$ and $\boldsymbol c=(0.5,0)$. The base state is $u_b=0$, and $\Lin$ is exact in Fourier space. The quadratic reaction saturates the response, so that at $T$ a blob of unit intensity departs from the linear response by four fifths, and training uses $I_s\le2$. The problem is posed on the periodic square $[0,2\pi)^2$ with a $48\times48$ grid ($N=2304$) and a horizon $T=1.5$ divided into $K=30$ intervals, and nine point probes on a $3\times3$ lattice observe it. The source is a Gaussian with position $(x_s,y_s)$, intensity $I_s$ and width $\sigma_s\in[0.35,0.6]$. The encoder is a periodic two-dimensional dilated convolutional network with tanh activations and width 16. The ablations use $c=2$ channels (lifted dimension $d=6{,}912$). Table~\ref{tab:main}, Fig.~\ref{fig:kolmores} and the reuse study use $c=4$, and Fig.~\ref{fig:rdres}, including both the posterior and the inversion sweep, uses the eight-channel equivariant lift of Appendix~\ref{app:equivariant}. The two-channel and four-channel lifts are trained on 400 random members of a one-or-two-blob family. In this family a single blob carries the whole intensity budget or, with probability one half, two blobs share it, so the same lift serves both single-source and multi-source inference.

\textbf{Sheared advection: two-dimensional Kolmogorov flow.} The dynamics are the Navier--Stokes equations in vorticity form, $\omega_t+\boldsymbol u\cdot\nabla\omega=\nu\nabla^2\omega+f$, with forcing $f=-n_f\cos n_fy$, $n_f=4$ and $\nu=0.1$. We perturb them about the steady laminar base $\omega_b=f/(\nu n_f^2)$ on the same periodic square, grid, horizon and $3\times3$ probe lattice as reaction--diffusion. This base is weakly unstable at this viscosity, but its fastest-growing linear mode grows by only $0.2\%$ over the horizon. The solver is pseudo-spectral with $2/3$ dealiasing and uses a Strang-split step with fourth-order Runge--Kutta substeps, which is second order overall. We form the propagator $\Lin$ by central differences of $N$ batched perturbations with a step of $10^{-6}$. Because the base is steady, $\Lin$ is the same at every step, but unlike in reaction--diffusion it is not diagonal in Fourier space. The source is a Gaussian injection of vorticity with the same parameters as in reaction--diffusion, and the finite-amplitude effect is the self-advection of the injected vortex within the shear. Training uses $I_s\le4$ on 400 members of the same one-or-two-blob family, and the encoder and channel counts are those of reaction--diffusion.

\subsection{Inference protocol, training and sampling}
\textbf{Inference protocol.} The training amplitude $I_{\max}$ is the largest source intensity in the training family, which is $2$ on reaction--diffusion, $4$ on Kolmogorov flow and $0.7$ on Burgers. The measurement error of a band is $\varepsilon_m=\sum\|\hat{\boldsymbol m}-\boldsymbol m\|/\sum\|\boldsymbol m\|$ over the sources in that band. Table~\ref{tab:main} draws its test intensities from $[0.1\,I_{\max},I_{\max}]$, whereas the ablation tables draw them from $[0,I_{\max}]$, which is why their medians differ. Every probe carries independent Gaussian noise whose standard deviation is $1\%$ of the root-mean-square measurement over the test draw. The one-shot inversion picks the candidate nearest to the observation in measurement space, on a lattice of $24\times24$ positions (every second grid point) and $24$ intensities evenly spaced from $0.05$ to $I_{\max}$. The source width is known and fixed at $0.45$, although training draws it from $[0.35,0.6]$.

\textbf{Training and hardware.} We train the encoder with Adam at a learning rate of $10^{-3}$ on a cosine schedule for 300 iterations, raising the rollout length from $K/3$ to $2K/3$ and then to $K$. The ridge parameter in \eqref{eq:fit} is $\lambda=10^{-5}$ relative to the trace of the Gram matrix. The two-dimensional lifts were trained on a single RTX 3090 GPU. The encoder runs in single precision, the Gram products and the ridge solve run in double precision, and the trained encoder is cast to double precision for evaluation. The scripts in the supplementary code generate every table entry from the saved result files.

\textbf{Centering in closed form.} For a tanh network, both corrections in \eqref{eq:centering} have closed forms. The offset $\rawfeat_\theta(\boldsymbol 0)$ is obtained by passing the biases through the activations layer by layer, and $\boldsymbol J_0$ is the product of the layer weight matrices, each scaled by $\tanh'$ evaluated at the corresponding activation for zero input.

\textbf{Posterior sampling.} We sample the joint posterior of two sources by differential-evolution Markov chain Monte Carlo with 96 chains. The chains start from a Levenberg--Marquardt refinement of the best lattice pairs under the same model. During a burn-in of 800 steps the likelihood is annealed from $2\%$ of its strength to its full strength, and the following 1600 steps are retained. Because the two sources are exchangeable, we relabel the chains at every step, so that proposals are not wasted on differences between chains that label the sources in opposite orders. The prior is uniform over the periodic domain for each position and over $[0,I_{\max}]$ for each intensity. For the coverage statistics, the two sources of each sample are labelled so as to minimize their summed periodic distance to the true pair. Two independent sampler seeds agree on the posterior of a test pair to within its standard deviation, and the lifted functional is evaluated with the encoder in single precision.

\subsection{Translation-equivariant learned blocks}
\label{app:equivariant}
On a periodic domain with a uniform base state, as in reaction--diffusion, the solver, the anchor $\Lin$ and the convolutional encoder are all translation equivariant. Dense learned blocks are not, so they spend capacity relearning that a source at one place behaves like the same source elsewhere. A translation-equivariant linear map is a convolution, so imposing this symmetry makes each learned block diagonal in wavenumber. This shrinks the feature block by three orders of magnitude, splits the fit into one $c\times c$ system per wavenumber, and makes stability easy to enforce, because capping the spectral radius becomes $N$ independent $c\times c$ problems. On reaction--diffusion, whose linearized propagator is a Fourier multiplier, the equivariant blocks reduce the forward error to $0.19\%$ with four channels and $0.05\%$ with sixteen, compared with $116\%$ for the linearized model.

\section{Structural evidence and inference results}
\label{app:tables}
This appendix gives the full results for each problem. It first presents the structural evidence that the constraints and the lift deliver what Section~\ref{sec:structure} promises, and then the inference of one source, of two sources, and with a single lift reused across sensor layouts and horizons.

\subsection{Structural evidence}
\label{app:structure}
This part isolates what the constraints and the lift contribute before any inference is run, namely the small-amplitude limit, the forward accuracy in each intensity band and the measurement gradient.

\begin{table}[h!]
\centering\footnotesize
\caption{The constrained lift is the most accurate model at small amplitude on all three systems, and it infers the intensity nearly as well as brute force. The columns give the measurement error $\varepsilon_m$ over the training range and at $I=10^{-3}$ for our lift, the unconstrained lift and the linear embedding, followed by the median intensity error of the inference. The two-dimensional rows use the two-channel lifts of Tables~\ref{tab:rd} and~\ref{tab:kolmo} on their own draw, so they differ from the four-channel values in Table~\ref{tab:main}. The unconstrained lift was not trained on Burgers. Bold marks the best value in each group, excluding brute force.}
\label{tab:summary}
\resizebox{\linewidth}{!}{\begin{tabular}{@{}l ccc ccc ccc@{}}\toprule
 & \multicolumn{3}{c}{$\varepsilon_m$ (\%), training range} & \multicolumn{3}{c}{$\varepsilon_m$ at $I=10^{-3}$} & \multicolumn{3}{c}{inference $\delta_I$ (\%)} \\
\cmidrule(lr){2-4}\cmidrule(lr){5-7}\cmidrule(lr){8-10}
problem & ours & unconstr. & linear & ours & unconstr. & linear & ours & brute force & linear \\ \midrule
Burgers, base flow & \textbf{0.9} & -- & 40.7 & {\boldmath$1.0\times10^{-4}$} & -- & $1.1\times10^{-3}$ & \textbf{5.8} & 5.8 & 20.1 \\
reaction--diffusion & 3.0 & \textbf{1.9} & 76.7 & {\boldmath$8.4\times10^{-5}$} & $2.3\times10^{-1}$ & $6.1\times10^{-4}$ & \textbf{12.6} & 10.7 & 35.9 \\
Kolmogorov flow & \textbf{5.7} & 9.7 & 18.4 & {\boldmath$9.0\times10^{-5}$} & $6.7\times10^{-1}$ & $2.3\times10^{-4}$ & \textbf{3.8} & 2.7 & 8.3 \\
\bottomrule\end{tabular}
}
\end{table}

\textbf{Summary.} Table~\ref{tab:summary} gathers the three systems. On each of them, the constrained lift predicts the measurement of a source of intensity $10^{-3}$ more accurately than the linear embedding, whereas the unconstrained lift is off by tens of per cent, and its median intensity error in the inversion is within two percentage points of brute force.

\begin{figure}[t]
\centering
\includegraphics[width=\linewidth]{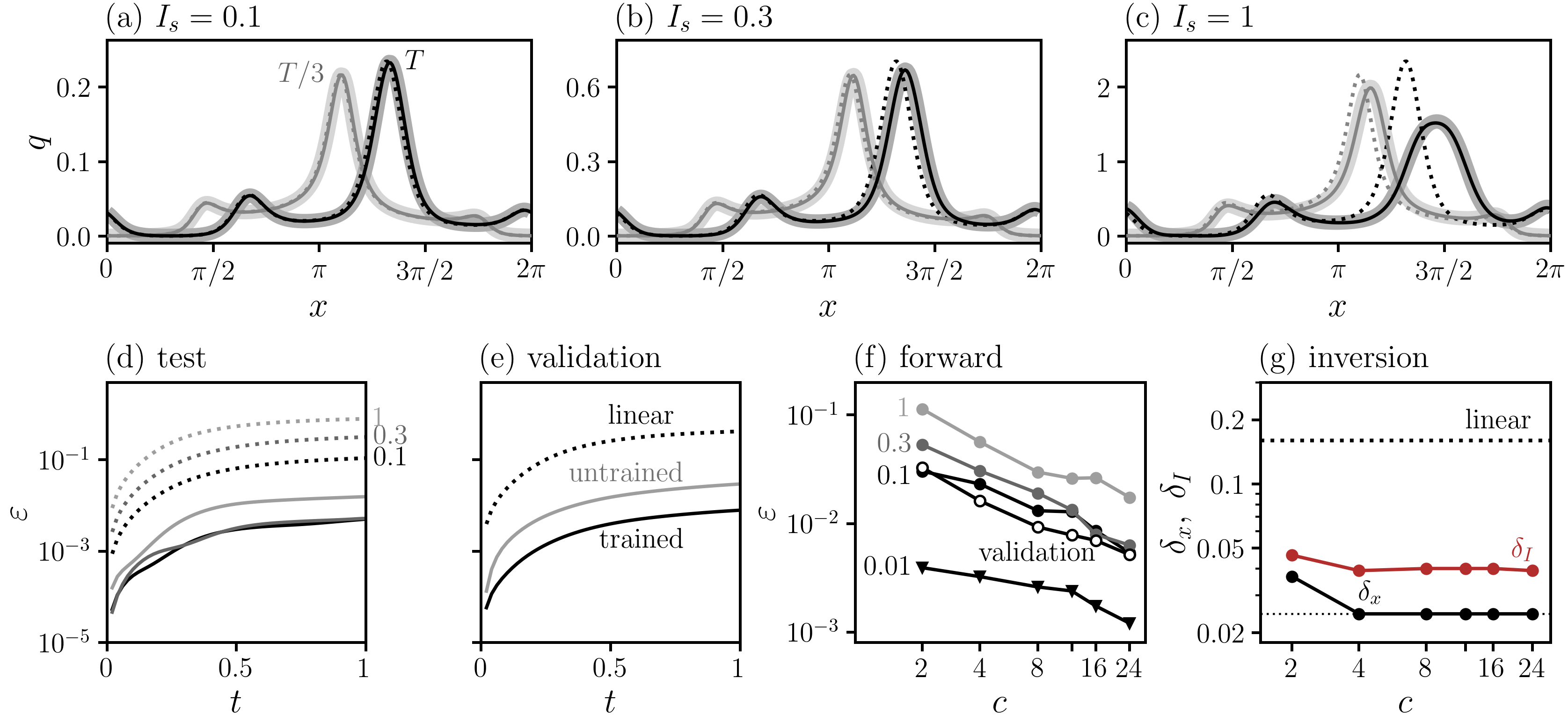}
\caption{Burgers about the base flow $1+\sin3x$. The lifted model follows the solver where the linearized solver does not. Its forward error falls steadily as the number of learned channels $c$ grows, whereas its inversion error stops improving at four channels. \textbf{(a--c)} The perturbation released at $x_s=3$ with the three test intensities, at $t=T/3$ (light grey) and at $T$ (black). The solver is the broad grey band, the lifted model the solid line and the linearized solver the dotted line. \textbf{(d)} Relative error over time at the three test intensities for the lifted model (solid) and the linearized solver (dotted), labelled by $I_s$. \textbf{(e)} Mean validation error over time for the linearized solver and for the lift before and after training, on the same vertical axis as (d). \textbf{(f)} Forward error at $T$ against $c$ for four test intensities, labelled by $I_s$, together with the mean validation error (open circles). \textbf{(g)} Median inversion error in position $\delta_x$ (black) and intensity $\delta_I$ (red) against $c$. The dotted line marked ``linear'' is the linear embedding, whose two errors differ by less than one per cent and are drawn as one line, and the fine dotted line marks one grid cell.}
\label{fig:burgers}
\end{figure}

\textbf{Burgers.} Table~\ref{tab:burgers} and Fig.~\ref{fig:burgers} give the results in each band on Burgers for a separate draw of 60 sources. The lifted model keeps the forward error at $T$ below $2\%$, whereas the error of the linearized solver reaches four fifths of the response. Its one-shot inversion matches brute force through the solver in intensity in every band and is at least as accurate in position. Figure~\ref{fig:burgers}f--g shows how the results depend on the number of channels.

\begin{table}[h!]
\centering\small
\caption{On Burgers the lifted model matches brute force in every band, while the errors of the linear embedding grow with the intensity. Top, relative error at $T$ of the forward models. Bottom, median position error (in length units, with grid cell $0.025$) and relative intensity error of the one-shot inversion in each intensity band (60 sources, $1\%$ noise), on a draw separate from the 800 sources of Table~\ref{tab:main}.}
\label{tab:burgers}
\begin{tabular}{lccc}\toprule
 & $I_s=0.1$ & $I_s=0.3$ & $I_s=1$ \\ \midrule
forward error at $T$, linearized solver & 10.9\% & 31.7\% & 79.4\% \\
forward error at $T$, lifted model & 0.5\% & 0.5\% & 1.6\% \\ \bottomrule
\end{tabular}
\quad
\begin{tabular}{lccc}\toprule
$I_s$ band & lifted & linear & brute force \\ \midrule
$[0.02,0.1)$ & 0.025 / 8.9\% & 0.074 / 9.3\% & 0.049 / 8.9\% \\
$[0.1,0.3)$ & 0.012 / 2.5\% & 0.123 / 16.4\% & 0.025 / 2.5\% \\
$[0.3,0.6)$ & 0.000 / 3.4\% & 0.295 / 17.1\% & 0.000 / 3.4\% \\
$[0.6,1.2)$ & 0.000 / 1.1\% & 0.491 / 32.4\% & 0.000 / 1.1\% \\ \bottomrule
\end{tabular}
\end{table}

\begin{figure}[t]
\centering\includegraphics[width=\linewidth]{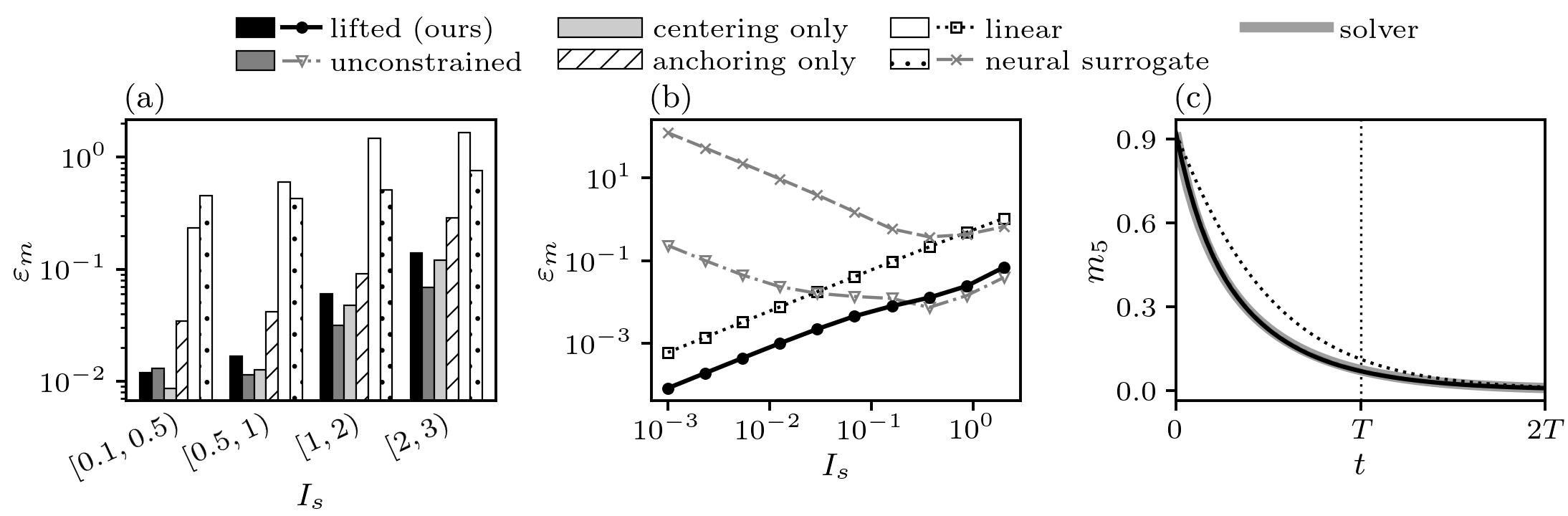}
\caption{On reaction--diffusion the measurement error of the full lift falls in proportion to the source intensity, whereas that of the unconstrained lift levels off at small amplitude. \textbf{(a)} Measurement error in each intensity band for the six models. \textbf{(b)} Measurement error against the source intensity. \textbf{(c)} The measurement of one probe over twice the training horizon for a source of intensity $1.5$, from the solver, the lifted functional, the linear embedding and the unconstrained lift.}
\label{fig:rd}
\end{figure}

\begin{figure}[t]
\centering\includegraphics[width=\linewidth]{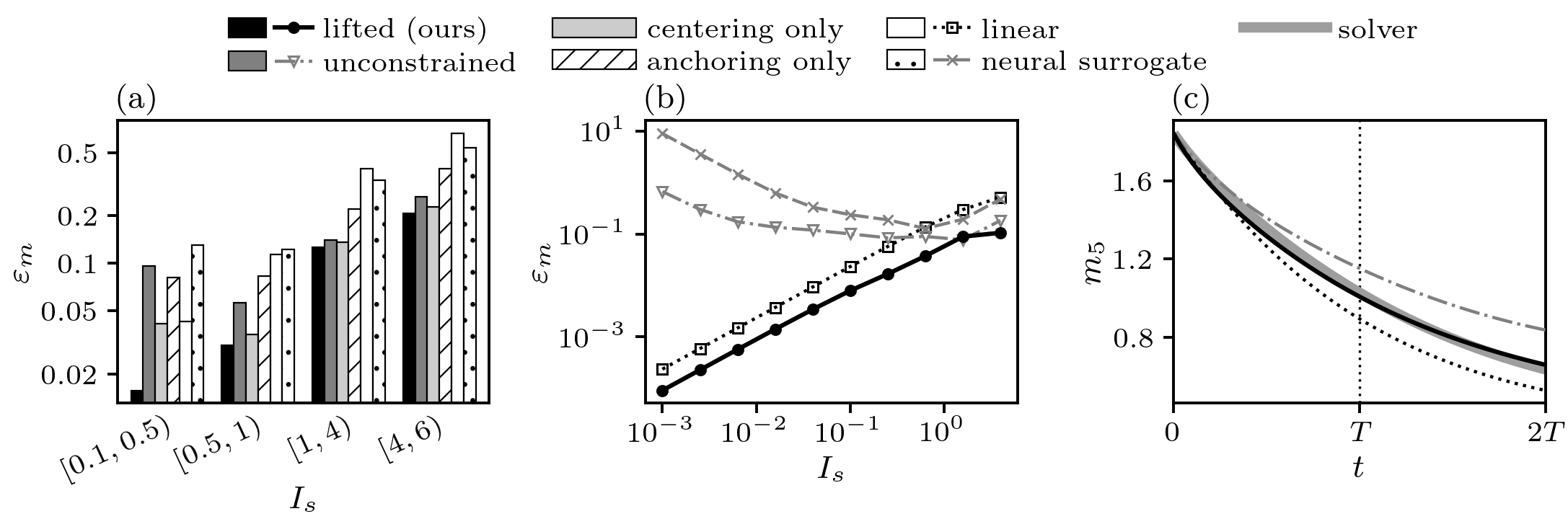}
\caption{On Kolmogorov flow the full lift is the most accurate model in every intensity band, and its measurement error falls with the source intensity. Panels as in Fig.~\ref{fig:rd}, with an injection of intensity $3$ in \textbf{(c)}.}
\label{fig:kolmo}
\end{figure}

\textbf{Ablation of the constraints.} Tables~\ref{tab:rd} and~\ref{tab:kolmo} remove the constraints one at a time at two channels, the size at which every variant was trained. Only the full lift, which is both anchored and centered, has a measurement error below $10^{-4}$ at $I_s=10^{-3}$ on both systems, which matches or improves on the accuracy of the linear adjoint. Centering alone leaves an error of a few per cent at that amplitude, and anchoring alone leaves an error larger than the measurement itself. At finite amplitude the full lift is the most accurate model in every band on Kolmogorov flow. On reaction--diffusion the unconstrained lift fits the two strongest bands about twice as closely, and the constraints trade this for the exact small-amplitude limit. Figures~\ref{fig:rd} and~\ref{fig:kolmo} show the same comparison as curves, together with one probe followed over twice the training horizon.

\begin{table}[h!]
\centering\small
\caption{On reaction--diffusion only the full lift reduces to the linear adjoint at small amplitude, while the unconstrained lift fits the strongest bands more closely. The columns give the measurement error in each intensity band, with the last band beyond the training amplitudes, and at $I_s=10^{-3}$, followed by the median position error in length units (candidate lattice $0.26$) and the relative intensity error, over 60 sources with $1\%$ noise and two channels. The neural surrogate did not converge on this problem and is reported as trained. Bold marks the best value in each column, excluding brute force.}
\label{tab:rd}
\begingroup\setlength{\tabcolsep}{4pt}\begin{tabular}{@{}l cccc c cc@{}}\toprule
 & \multicolumn{4}{c}{$\varepsilon_m$ (\%) per intensity band $I_s$} & $\varepsilon_m$ (ratio) at &  \multicolumn{2}{c}{inference} \\
\cmidrule(lr){2-5}\cmidrule(lr){7-8}
model & $[0.1,0.5)$ & $[0.5,1)$ & $[1,2)$ & $[2,3)$ & $I_s=10^{-3}$ & $\delta_x$ & $\delta_I$ (\%) \\ \midrule
lifted (ours) & 1.2 & 1.7 & 6.0 & 13.9 & {\boldmath$8.4\times10^{-5}$} & 0.145 & 12.6 \\
unconstrained lift & 1.3 & \textbf{1.1} & \textbf{3.2} & \textbf{6.9} & $2.3\times10^{-1}$ & \textbf{0.143} & \textbf{11.3} \\
centering only & \textbf{0.9} & 1.3 & 4.8 & 12.1 & $2.8\times10^{-2}$ & 0.155 & 15.7 \\
anchoring only & 3.4 & 4.2 & 9.2 & 28.8 & $1.9\times10^{0}$ & 0.181 & 14.3 \\
linear adjoint & 23.3 & 59.8 & 146.9 & 165.2 & $6.1\times10^{-4}$ & 0.147 & 35.9 \\
neural surrogate & 45.7 & 42.9 & 51.1 & 75.6 & $1.2\times10^{2}$ & 0.510 & 27.8 \\
\midrule brute force (reference) & -- & -- & -- & -- & -- & 0.124 & 10.7 \\ \bottomrule\end{tabular}\endgroup

\end{table}

\begin{table}[h!]
\centering\small
\caption{On Kolmogorov flow the full lift is the most accurate learned model in every band and at small amplitude. Columns as in Table~\ref{tab:rd}.}
\label{tab:kolmo}
\begingroup\setlength{\tabcolsep}{4pt}\begin{tabular}{@{}l cccc c cc@{}}\toprule
 & \multicolumn{4}{c}{$\varepsilon_m$ (\%) per intensity band $I_s$} & $\varepsilon_m$ (ratio) at &  \multicolumn{2}{c}{inference} \\
\cmidrule(lr){2-5}\cmidrule(lr){7-8}
model & $[0.1,0.5)$ & $[0.5,1)$ & $[1,4)$ & $[4,6)$ & $I_s=10^{-3}$ & $\delta_x$ & $\delta_I$ (\%) \\ \midrule
lifted (ours) & \textbf{1.6} & \textbf{3.0} & \textbf{12.6} & \textbf{20.7} & {\boldmath$9.0\times10^{-5}$} & 0.111 & \textbf{3.8} \\
unconstrained lift & 9.6 & 5.6 & 14.0 & 26.3 & $6.7\times10^{-1}$ & 0.114 & 4.7 \\
centering only & 4.1 & 3.5 & 13.5 & 22.6 & $5.8\times10^{-2}$ & \textbf{0.106} & 4.2 \\
anchoring only & 8.1 & 8.3 & 21.9 & 39.6 & $8.2\times10^{0}$ & 0.133 & 5.0 \\
linear adjoint & 4.3 & 11.3 & 39.6 & 65.9 & $2.3\times10^{-4}$ & 0.204 & 8.3 \\
neural surrogate & 13.0 & 12.2 & 33.5 & 53.6 & $9.0\times10^{0}$ & 0.183 & 7.6 \\
\midrule brute force (reference) & -- & -- & -- & -- & -- & 0.106 & 2.7 \\ \bottomrule\end{tabular}\endgroup

\end{table}

\textbf{Channel count.} Table~\ref{tab:channels} varies the number of channels of the full lift. Going from two to four channels reduces the measurement error in the top band from $6.0\%$ to $1.6\%$ on reaction--diffusion and from $12.6\%$ to $7.6\%$ on Kolmogorov flow, while keeping the small-amplitude limit and improving the inference. This is why Table~\ref{tab:main} and the inference studies use four channels.

\begin{table}[h!]
\centering\small
\caption{With four channels the anchored, centered lift is the most accurate of the three sizes on both two-dimensional systems. The columns give the feature dimension $d_\Phi=cN$, the number of training iterations, the best validation rollout MSE, the measurement error in the top training band and beyond it, the error at $I_s=10^{-3}$, and the median inference errors. Every row uses the same schedule and training pairs, although the cost of the fit grows with $c$. We vary $c$ only for the anchored lift, because the full-propagator fits of the other variants exceed the GPU memory at $c=4$. Bold marks the best value in each column.}
\label{tab:channels}
\begingroup\setlength{\tabcolsep}{4pt}\begin{tabular}{@{}rrr c cc c cc@{}}\toprule
 & & & val. & \multicolumn{2}{c}{$\varepsilon_m$ (\%)} & $\varepsilon_m$ (ratio) at & \multicolumn{2}{c}{inference} \\
\cmidrule(lr){5-6}\cmidrule(lr){8-9}
$c$ & $d_\Phi$ & iters & MSE & top band & beyond & $I_s=10^{-3}$ & $\delta_x$ & $\delta_I$ (\%) \\ \midrule
\multicolumn{9}{@{}l}{\emph{reaction--diffusion}} \\
1 & 2{,}304 & 300 & $1.7\times10^{-3}$ & 11.4 & 9.0 & $9.3\times10^{-5}$ & 0.192 & 13.7 \\
2 & 4{,}608 & 300 & $7.8\times10^{-4}$ & 6.0 & 13.9 & $8.4\times10^{-5}$ & 0.145 & 12.6 \\
4 & 9{,}216 & 300 & {\boldmath$7.2\times10^{-5}$} & \textbf{1.6} & \textbf{4.0} & {\boldmath$8.3\times10^{-6}$} & \textbf{0.113} & \textbf{8.7} \\
\addlinespace\multicolumn{9}{@{}l}{\emph{Kolmogorov flow}} \\
1 & 2{,}304 & 300 & $5.5\times10^{-3}$ & 14.4 & 25.4 & $2.9\times10^{-4}$ & \textbf{0.106} & 4.3 \\
2 & 4{,}608 & 300 & $2.9\times10^{-3}$ & 12.6 & 20.7 & $9.0\times10^{-5}$ & 0.111 & 3.8 \\
4 & 9{,}216 & 300 & {\boldmath$1.3\times10^{-3}$} & \textbf{7.6} & \textbf{16.2} & {\boldmath$6.7\times10^{-5}$} & 0.106 & \textbf{3.7} \\
\bottomrule\end{tabular}\endgroup

\end{table}

\textbf{Measurement gradient.} Table~\ref{tab:gradcheck} compares the gradient of the predicted measurements with respect to the source parameters against the solver's. For the weakest sources the constrained lift gives the most accurate gradient among the learned models on both systems, as the constraints require. At finite amplitude the ranking follows the forward errors of Tables~\ref{tab:rd} and~\ref{tab:kolmo}, with the unconstrained lift more accurate on reaction--diffusion and the constrained lift more accurate on Kolmogorov flow.

\begin{table}[h!]
\centering\small
\caption{For the weakest sources the constrained lift gives the most accurate measurement gradient among the learned models. We difference every model and the solver in the same way, centrally, with steps of $10^{-3}$ in position and $10^{-3}$ (relative) in intensity, and reducing the steps threefold changes no entry. Each entry is the median, over twelve sources in the band, of $\|\boldsymbol J-\boldsymbol J_{\mathrm{solver}}\|_F/\|\boldsymbol J_{\mathrm{solver}}\|_F$ in per cent, where $\boldsymbol J$ is the Jacobian of the nine probe measurements with respect to $(x_s,y_s,I_s)$, made dimensionless by the source width and intensity. The models are those of Tables~\ref{tab:rd} and~\ref{tab:kolmo}, and the last band of each system lies beyond the training range. Bold marks the most accurate model in each band.}
\label{tab:gradcheck}
\resizebox{\linewidth}{!}{\begingroup\setlength{\tabcolsep}{3pt}\begin{tabular}{@{}l ccccc ccccc@{}}\toprule
 & \multicolumn{5}{c}{reaction--diffusion, $I_s$ band} & \multicolumn{5}{c}{Kolmogorov flow, $I_s$ band} \\
\cmidrule(lr){2-6}\cmidrule(lr){7-11}
model & $[0.01,0.1)$ & $[0.1,0.5)$ & $[0.5,1)$ & $[1,2)$ & $[2,3)$ & $[0.01,0.1)$ & $[0.1,0.5)$ & $[0.5,1)$ & $[1,4)$ & $[4,6)$ \\ \midrule
lifted (ours) & \textbf{0.7} & 3.3 & 7.5 & 21 & 73 & 0.6 & 2.6 & 7.9 & \textbf{17} & 35 \\
unconstrained lift & 4.9 & \textbf{2.9} & \textbf{4.7} & \textbf{7.8} & \textbf{24} & 16 & 13 & 18 & 25 & 37 \\
centering only & 2.9 & 3.1 & 5.3 & 17 & 60 & 10 & 8.1 & 11 & 20 & \textbf{29} \\
anchoring only & 11 & 7.0 & 12 & 23 & 114 & 17 & 19 & 19 & 27 & 54 \\
neural surrogate & 66 & 61 & 66 & 94 & 170 & 18 & 17 & 17 & 37 & 68 \\
linear embedding & 5.1 & 28 & 65 & 205 & 387 & 1.1 & 4.7 & 10 & 43 & 73 \\
quadratic correction & 1.5 & 3.5 & 7.9 & 179 & 716 & \textbf{0.0} & \textbf{0.3} & \textbf{1.9} & 18 & 66 \\
\bottomrule\end{tabular}\endgroup
}
\end{table}

\subsection{Single-source inference}
This part shows how the lifted functional locates a single source on the two-dimensional systems, and how its estimates compare with brute force through the solver.

\begin{figure}[t]
\centering\includegraphics[width=\linewidth]{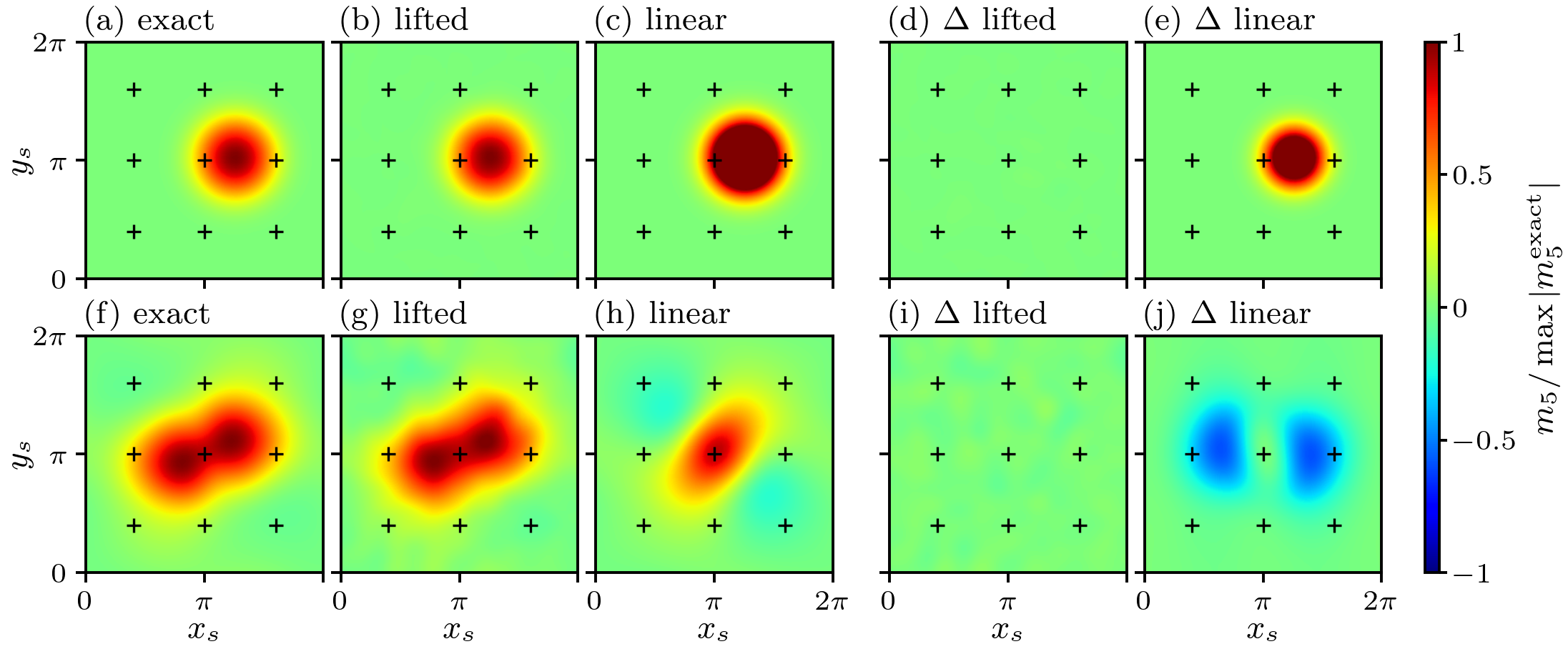}
\caption{The lifted functional reproduces the finite-amplitude sensitivity map of the centre probe at the cost of one inner product per source position, whereas the linear embedding misses both its amplitude and its shape. Reaction--diffusion is shown in \textbf{(a--e)} and Kolmogorov flow in \textbf{(f--j)}, at $I_s=0.75\,I_{\max}$ with the four-channel lifts. Each row shows the exact map, which costs one nonlinear solve per source position, the lifted functional and the linear embedding, followed by the difference of each approximation from the exact map. Every panel in a row is divided by the largest value of that row's exact map, and all ten panels share one colour scale, so the differences read directly as fractions of the exact map. Values beyond $\pm1$ saturate, and the crosses mark the probes.}
\label{fig:sens}
\end{figure}

\textbf{Sensitivity maps.} Figure~\ref{fig:sens} shows the object that the inference searches, namely the measurement of one probe as a function of the source position at a fixed finite intensity. The exact map costs one nonlinear solve per position, whereas the lifted functional evaluates it with one inner product per position.

\begin{figure}[t]
\centering\includegraphics[width=\linewidth]{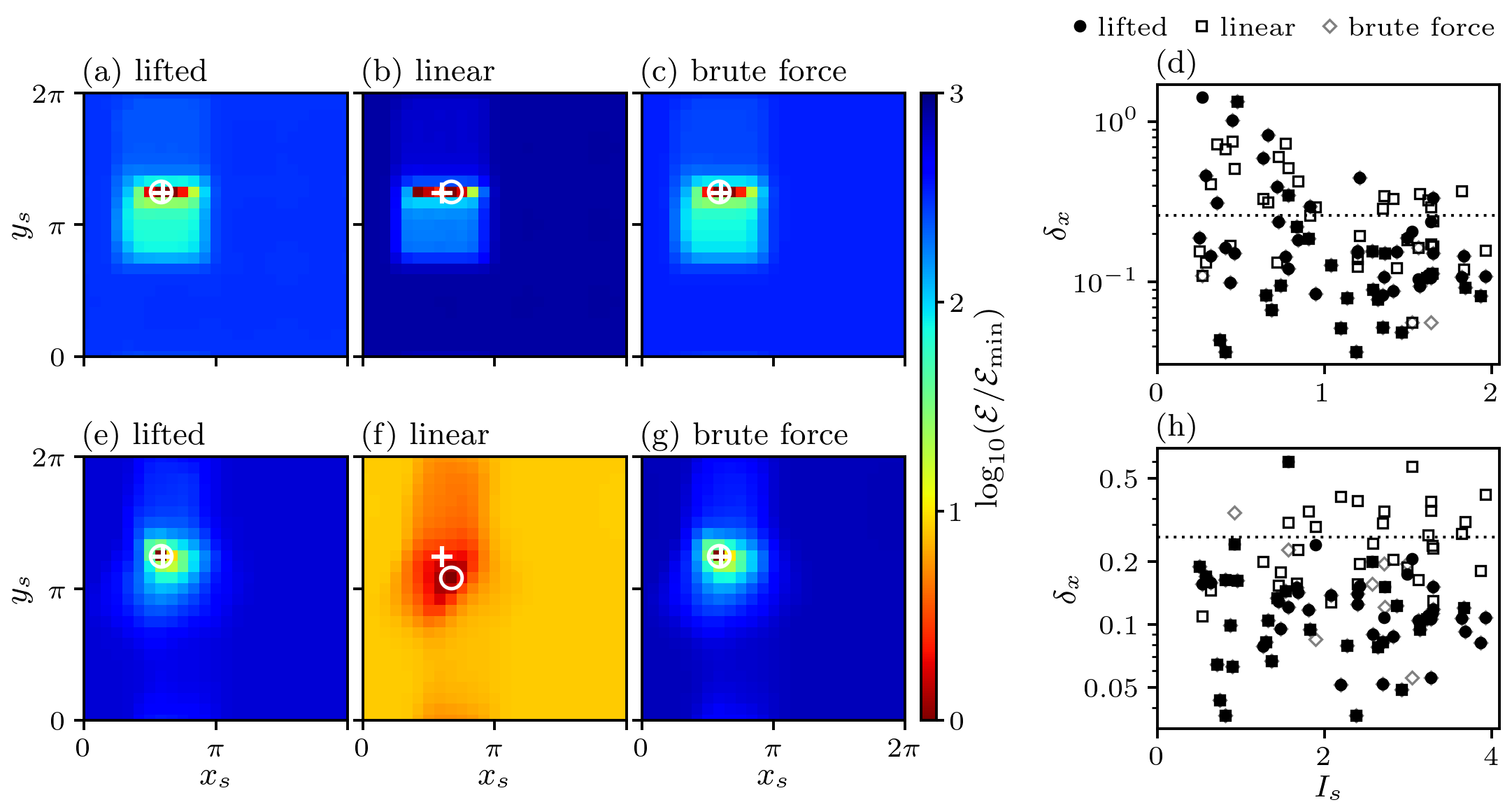}
\caption{The misfit landscape of the lifted functional has its minimum at the true source, as for brute force through the solver, whereas the minimum of the linear embedding is displaced on Kolmogorov flow. The panels show single-source inference with the four-channel lifts, for reaction--diffusion in \textbf{(a--d)} and Kolmogorov flow in \textbf{(e--h)}. The first three panels of each row show the misfit over the candidate lattice for the lifted functional, the linear embedding and brute force through the solver. At each position the misfit is minimized over the intensity, normalized by its own minimum and drawn on one logarithmic scale, on which red indicates a good fit. The cross marks the true source and the circle the estimate. Panels (d) and (h) show the position error of the 60 test sources against their intensity, with the candidate-lattice spacing dotted.}
\label{fig:inference}
\end{figure}

\textbf{Misfit landscapes and uncertainty.} Figure~\ref{fig:inference} shows the misfit over the candidate lattice for one test source, and the position error of all 60 test sources against their intensity. Table~\ref{tab:ci} gives $95\%$ bootstrap intervals for the medians in Table~\ref{tab:main}. The paired interval for the difference between the lifted model and the solver contains zero for position on all three systems, and for intensity on Burgers and reaction--diffusion. On Kolmogorov flow the lifted model is less accurate in intensity by $0.2$ to $2.1$ percentage points.

\begin{table}[h!]
\centering\small
\caption{The lifted inference is statistically indistinguishable from the solver in position on all three systems. Medians of Table~\ref{tab:main} with $95\%$ bootstrap intervals over the test sources ($10{,}000$ resamples).}
\label{tab:ci}
\begin{tabular}{@{}ll cc@{}}\toprule
case & model & $\delta_x$ (cells) & $\delta_I$ (\%) \\ \midrule
\multirow{4}{*}{Burgers} & linear & 6.00 [5.00, 6.00] & 15.2 [13.8, 16.5] \\
 & quadratic & 3.00 [2.00, 3.00] & 8.9 [8.2, 9.7] \\
 & lifted & 1.00 [1.00, 1.00] & 3.6 [3.3, 3.9] \\
 & solver & 1.00 [1.00, 1.00] & 3.4 [3.1, 3.7] \\
\midrule
\multirow{4}{*}{reaction--diffusion} & linear & 0.63 [0.51, 0.84] & 40.2 [34.7, 45.5] \\
 & quadratic & 0.91 [0.60, 1.21] & 37.1 [23.2, 44.6] \\
 & lifted & 0.55 [0.41, 0.60] & 14.2 [10.8, 18.2] \\
 & solver & 0.46 [0.41, 0.58] & 14.4 [10.8, 17.2] \\
\midrule
\multirow{4}{*}{Kolmogorov flow} & linear & 0.63 [0.51, 0.74] & 7.2 [4.6, 9.6] \\
 & quadratic & 0.45 [0.40, 0.54] & 6.3 [4.5, 8.2] \\
 & lifted & 0.41 [0.37, 0.48] & 3.9 [3.0, 4.8] \\
 & solver & 0.41 [0.36, 0.46] & 2.8 [2.1, 3.8] \\
\bottomrule\end{tabular}

\end{table}

\subsection{Two sources}
\label{app:twosource}
This part reports the two-source study for every pair family on both two-dimensional systems. We use three families. In the first, the two sources are more than $4\sigma_s$ apart with $0.35$--$0.5\,I_{\max}$ each. In the second, strong family they are also separated but carry $0.5$--$1.0\,I_{\max}$ each, so the combined field reaches up to twice the training budget. In the third they overlap, at $1$--$3\sigma_s$ apart. Each family has 30 pairs on reaction--diffusion and 24 on Kolmogorov flow, read by a $4\times4$ probe lattice with $1\%$ noise. Every model scores all pairs of single-source candidates under superposition, and the best lattice pairs are then refined by L-BFGS on the model's own misfit, built from the duality relation \eqref{eq:family}. The study uses the four-channel dense lifts of Table~\ref{tab:channels} rather than the eight-channel equivariant lift of Section~\ref{sec:inference}, which makes the comparison against the linear embedding a conservative one.

\begin{figure}[t]
\centering\includegraphics[width=\linewidth]{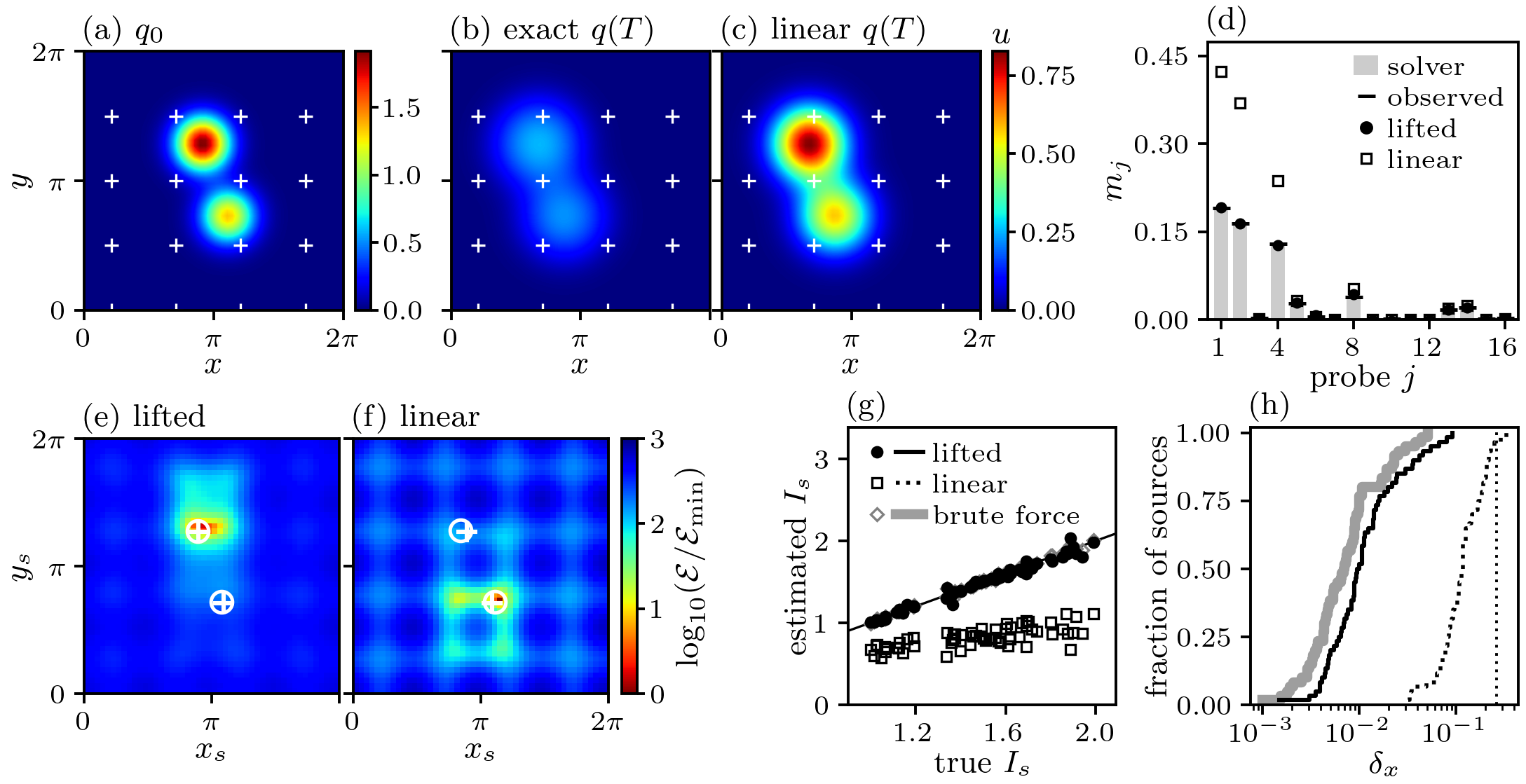}
\caption{The lifted functional recovers both sources of a separated pair on reaction--diffusion, together with their intensities, as accurately as brute force, whereas the linear embedding underestimates every intensity. The pairs carry $0.5$--$1.0\,I_{\max}$ per source, up to twice the training budget in total, and the frame is shifted to centre the pair shown. \textbf{(a--d)} The forward problem. Panel (a) shows the initial field with the $4\times4$ probes as crosses, (b) the exact field at $T$, and (c) the superposition of the linearized responses on the same colour scale as (b). Panel (d) shows the sixteen measurements, from the solver as bars and with $1\%$ noise as dashes, together with the predictions of the lifted functional and of the linear embedding at the true parameters. \textbf{(e--h)} The inverse problem. Panels (e) and (f) show the misfit of the second source over the domain at its fitted intensity, given the fitted first source, for the lifted functional and for the linear embedding, with crosses at the true sources and circles at the estimates. Panel (g) shows the estimated against the true intensity of every source over all pairs, and (h) the cumulative distribution of the per-source position error, with the candidate-lattice spacing dotted.}
\label{fig:twosource}
\end{figure}

\begin{figure}[t]
\centering\includegraphics[width=\linewidth]{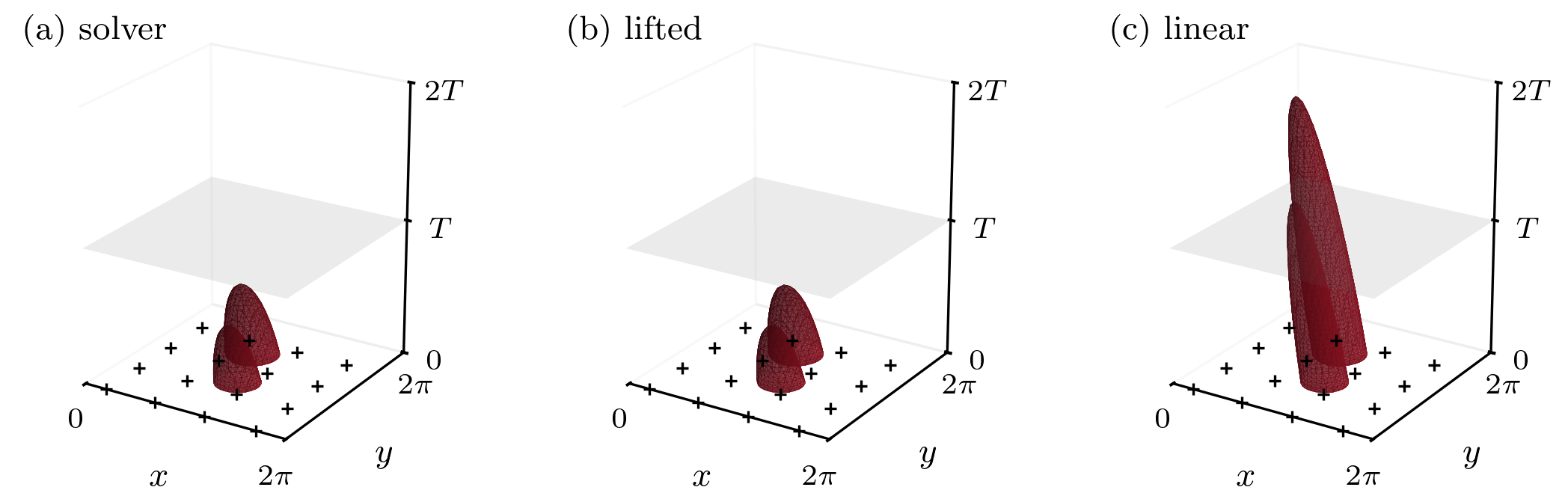}%
\caption{Beyond the training horizon, the lifted rollout reproduces the decay of the strong separated pair of Fig.~\ref{fig:twosource}, whereas the linear superposition stays twice as strong. The panels show isosurfaces of $u(x,y,t)$ at $25\%$ of the initial maximum, with time running vertically to $2T$, the translucent plane at the training horizon $T$ and the probes marked by crosses, for \textbf{(a)} the solver, \textbf{(b)} the lifted rollout and \textbf{(c)} the superposition of the linearized responses. Over the whole space--time field the lifted rollout is off by $3.5\%$ and the linear superposition by $81\%$.}
\label{fig:spacetime_rd}
\end{figure}

Table~\ref{tab:twosource} collects the results for all families. On reaction--diffusion the lifted functional places both sources within one lattice cell in $80$--$100\%$ of pairs, against $33$--$97\%$ for the linear embedding. It infers the intensities to within $10\%$, where the linear embedding misses them by $34$--$49\%$, and it runs 18 to 26 times faster than brute force. On Kolmogorov flow the two sources interact strongly, changing the measurement by $43$--$48\%$, and the lifted functional still resolves both sources more often than the linear embedding in every family. Figure~\ref{fig:twosource} shows one strong separated pair on reaction--diffusion, with the forward problem on top and the inverse problem below. Figure~\ref{fig:spacetime_rd} follows the same pair past the training horizon to $2T$, and Fig.~\ref{fig:kolmores}a--d in the main text shows the most strongly interacting pair on Kolmogorov flow.

\begin{table}[h!]
\centering\footnotesize
\caption{The lifted functional resolves two sources more often and more accurately than the linear embedding in every pair family. Interaction is the median departure of the exact response from the sum of the single-source responses. For each model the table gives the median per-source position error $\delta_x$ (lattice spacing $0.26$), the median intensity error $\delta_I$ and the fraction of pairs with both sources within one lattice cell, followed by the ratio of brute-force to lifted inference time. Bold marks the better of the lifted functional and the linear embedding.}
\label{tab:twosource}
\resizebox{\linewidth}{!}{\begin{tabular}{@{}ll c ccc ccc ccc c@{}}\toprule
 & & inter- & \multicolumn{3}{c}{lifted functional} & \multicolumn{3}{c}{linear embedding} & \multicolumn{3}{c}{brute force} & speed-up \\
\cmidrule(lr){4-6}\cmidrule(lr){7-9}\cmidrule(lr){10-12}
problem & pairs & action (\%) & $\delta_x$ & $\delta_I$ (\%) & both (\%) & $\delta_x$ & $\delta_I$ (\%) & both (\%) & $\delta_x$ & $\delta_I$ (\%) & both (\%) & brute/lifted \\ \midrule
reaction--diffusion & separated & 0.3 & \textbf{0.009} & \textbf{0.7} & \textbf{100.0} & 0.092 & 33.9 & 96.7 & 0.008 & 0.6 & 100.0 & 18 \\
 & separated, strong & 0.9 & \textbf{0.010} & \textbf{1.3} & \textbf{100.0} & 0.116 & 42.2 & 96.7 & 0.007 & 0.7 & 100.0 & 24 \\
 & overlapping & 16.1 & \textbf{0.035} & \textbf{9.0} & \textbf{83.3} & 0.145 & 37.1 & 63.3 & 0.027 & 6.4 & 93.3 & 24 \\
 & overlapping, strong & 22.6 & \textbf{0.045} & \textbf{9.4} & \textbf{80.0} & 0.208 & 49.4 & 33.3 & 0.024 & 5.8 & 100.0 & 26 \\
Kolmogorov flow & separated & 47.9 & \textbf{0.168} & \textbf{6.3} & \textbf{79.2} & 0.343 & 30.0 & 20.8 & 0.007 & 0.4 & 100.0 & 240 \\
 & separated, strong & 44.7 & \textbf{0.244} & \textbf{9.3} & \textbf{37.5} & 0.392 & 41.7 & 12.5 & 0.007 & 0.3 & 100.0 & 243 \\
 & overlapping & 43.1 & \textbf{0.283} & \textbf{46.2} & \textbf{20.8} & 0.519 & 47.0 & 4.2 & 0.037 & 4.3 & 70.8 & 248 \\
\bottomrule\end{tabular}
}
\end{table}

\subsection{Reuse across sensor layouts and horizons}
This part tests the claim that one trained lift serves any sensor layout and measurement time with a single backward solve and no new data. Table~\ref{tab:reuse} compares the four-channel lift with models trained directly on the same simulations, using the paper's $3\times3$ probes, two held-out layouts and a held-out horizon. The lift is more accurate than the direct CNN in every column on both systems. The parameter MLP is off by $13$--$130\%$ for the weakest sources and cannot represent pairs. The direct readout, a ridge regression on the lift's own features that is refitted for each layout and horizon, stays close to the lift throughout. This shows that the features the lift learns through the dynamics carry the finite-amplitude response.

\begin{table}[h!]
\centering\small
\caption{One lift serves new sensor layouts and horizons as accurately as models refitted for each of them. The table gives the relative measurement error (\%) of the four-channel lift and of models trained directly on the same simulations, for single sources and for overlapping pairs. The direct readout and the direct CNN both keep the exact linear term $\langle\qL,S\rangle$ and add a linear readout of centered features, so they are anchored and centered like the lift. The readout is a ridge regression on the lift's own features, fitted to the stored training fields at each layout and horizon. The direct CNN, which has the encoder's architecture but is trained on measurements alone, and the parameter MLP must be retrained for each layout.}
\label{tab:reuse}
\resizebox{\linewidth}{!}{\begingroup\setlength{\tabcolsep}{3pt}\begin{tabular}{@{}l ccccc ccccc@{}}\toprule
 & \multicolumn{5}{c}{reaction--diffusion} & \multicolumn{5}{c}{Kolmogorov flow} \\
\cmidrule(lr){2-6}\cmidrule(lr){7-11}
model & $3{\times}3$, $T$ & $4{\times}4$, $T$ & 25 rand., $T$ & $3{\times}3$, $T/2$ & pairs & $3{\times}3$, $T$ & $4{\times}4$, $T$ & 25 rand., $T$ & $3{\times}3$, $T/2$ & pairs \\
\midrule
\multicolumn{11}{@{}l}{\emph{top training band, $I_s\in[1,I_{\max})$}} \\
lifted (ours) & 1.5 & 1.6 & 1.5 & 1.4 & 0.8 & 6.0 & 11 & 5.8 & 4.4 & 5.2 \\
direct readout & 2.3 & 2.2 & 2.0 & 1.5 & 1.3 & 4.6 & 9.4 & 4.7 & 3.1 & 4.6 \\
direct CNN & 9.2 & 11 & -- & -- & 5.1 & 17 & 25 & -- & -- & 8.5 \\
parameter MLP & 5.8 & 5.5 & 5.3 & 7.8 & -- & 4.6 & 15 & 4.9 & 5.0 & -- \\
\midrule
\multicolumn{11}{@{}l}{\emph{weakest band, $I_s\in[0.1,0.5)$}} \\
lifted (ours) & 0.2 & 0.1 & 0.2 & 0.2 &  & 1.2 & 2.4 & 1.3 & 0.9 &  \\
direct readout & 0.5 & 0.4 & 0.3 & 0.3 &  & 1.1 & 2.6 & 1.2 & 0.9 &  \\
direct CNN & 2.6 & 2.1 & -- & -- &  & 1.6 & 3.8 & -- & -- &  \\
parameter MLP & 14 & 13 & 13 & 23 &  & 28 & 130 & 34 & 49 &  \\
\bottomrule\end{tabular}\endgroup
}
\end{table}

\textbf{Horizons beyond training and scarce data.} The lift was trained on trajectories that end at $T$. When we iterate its propagator to $1.5T$, its measurement error stays within $13\%$ on reaction--diffusion and $17\%$ on Kolmogorov flow in every intensity band, against up to $196\%$ and $63\%$ for the linear embedding, whereas a direct readout has no data at that horizon. With 50 instead of 400 training trajectories, the lift is up to four times more accurate than a readout fitted to the same trajectories on reaction--diffusion (Table~\ref{tab:efficiency}). Anchoring and centering alone do not explain this accuracy. An anchored and centered CNN with the lift's architecture, trained end to end on the same trajectories but without the propagator, is $5$ to $16$ times less accurate than the lift on reaction--diffusion, and less accurate in every band on Kolmogorov flow, with 400 trajectories.

\begin{table}[h!]
\centering\small
\caption{The lift predicts beyond its training horizon, where the linear embedding fails. The table gives the relative measurement error (\%) on the $3\times3$ probes in the four intensity bands, at the training horizon $T$ and at $1.5T$. It compares the four-channel lift trained on 400 and on 50 trajectories, a ridge readout fitted on each lift's own features and trajectories, an anchored and centered CNN of the lift's architecture trained end to end on the same trajectories, and the linear embedding. The readout and the CNN have no training data beyond $T$.}
\label{tab:efficiency}
\resizebox{\linewidth}{!}{\begingroup\setlength{\tabcolsep}{2.5pt}\begin{tabular}{@{}l cccc cccc cccc cccc@{}}\toprule
 & \multicolumn{8}{c}{reaction--diffusion} & \multicolumn{8}{c}{Kolmogorov flow} \\
\cmidrule(lr){2-9}\cmidrule(lr){10-17}
 & \multicolumn{4}{c}{$T$} & \multicolumn{4}{c}{$1.5T$} & \multicolumn{4}{c}{$T$} & \multicolumn{4}{c}{$1.5T$} \\
\cmidrule(lr){2-5}\cmidrule(lr){6-9}\cmidrule(lr){10-13}\cmidrule(lr){14-17}
band & 1 & 2 & 3 & 4 & 1 & 2 & 3 & 4 & 1 & 2 & 3 & 4 & 1 & 2 & 3 & 4 \\ \midrule
lifted, 400 traj. & 0.2 & 0.3 & 1.5 & 4.2 & 0.3 & 0.8 & 4.7 & 13 & 1.2 & 1.6 & 6.0 & 12 & 1.4 & 2.2 & 6.6 & 17 \\
readout, 400 traj. & 0.5 & 0.7 & 2.3 & 5.2 & -- & -- & -- & -- & 1.1 & 2.1 & 4.6 & 10 & -- & -- & -- & -- \\
anchored CNN, 400 traj. & 3.4 & 3.3 & 12 & 67 & -- & -- & -- & -- & 2.3 & 4.0 & 9.1 & 14 & -- & -- & -- & -- \\
lifted, 50 traj. & 3.9 & 8.8 & 32 & 84 & 6.2 & 15 & 56 & 140 & 2.8 & 5.5 & 17 & 38 & 3.1 & 6.5 & 20 & 41 \\
readout, 50 traj. & 16 & 31 & 52 & 96 & -- & -- & -- & -- & 2.1 & 4.3 & 15 & 28 & -- & -- & -- & -- \\
anchored CNN, 50 traj. & 18 & 39 & 67 & 161 & -- & -- & -- & -- & 1.9 & 4.8 & 18 & 39 & -- & -- & -- & -- \\
linear embedding & 27 & 70 & 106 & 169 & 31 & 86 & 134 & 196 & 3.2 & 7.5 & 32 & 57 & 3.5 & 8.3 & 35 & 63 \\
\bottomrule\end{tabular}\endgroup
}
\end{table}

\section{Burgers about a quiescent base: where the constraints came from}
\label{app:quiescent}
This appendix reports a Burgers study about a quiescent base state, $u_b=0$, in which the field is itself the perturbation and $\Lin$ is the exact heat propagator over one interval. It is not one of the three systems of Section~\ref{sec:experiments}. We keep it because the structural constraints were developed in this configuration, and because it contains the comparison with the unconstrained lift and with fitted dictionaries that motivates them. The horizon is $T=1.5$ in $K=75$ intervals, and the source width is $\sigma=0.5$. The test intensities are $I_s\in\{0.5,1.5,3\}$ at $x_s=2$, and training uses 800 sums of one to three Gaussians with amplitudes in $[0,2]$, so the strongest test lies beyond the training range. Every number below is read from the saved outputs listed in the results manifest of the repository.

\paragraph{Forward accuracy does not certify a sensitivity.} At equal size, with eight channels and 300 iterations, the unconstrained Koopman autoencoder and the anchored, centered lift are indistinguishable on the validation set, both with a mean rollout error of $1.3\%$. A Carleman lift fitted on twelve POD modes reaches $6.4\%$, and a local polynomial dictionary $[u,u^2,u^3]$ reaches $6.8\%$, while the physics-built Carleman generator diverges once $I_s\ge1.5$. The two neural lifts nevertheless differ in what they transport backward, and the difference shows up in that object alone, without reference to any adjoint taken at the perturbed state. When a sensor kernel of width $0.25$ at $x_m=4$ is propagated backward over the horizon through the transposed propagator, the unconstrained lift amplifies it by a factor of $217$ in norm, and its state channel alone by $6.1$, whereas the heat adjoint, to which that channel should reduce, decays. The anchored, centered lift keeps the same channel at $0.64$, which by (C2) and (C3) is exactly the linear adjoint. The backward field of the unconstrained model also carries oscillations at wavenumbers that the training data never excite, which is the signature of directions of the feature Gram matrix that the data leave unconstrained.

\paragraph{Small-amplitude limit.} The observables of the unconstrained lift do not vanish at the zero field, where their norm is of order $\mathcal{O}(10)$ in the trained models, and its fitted propagator cancels this offset only up to a bias of $10^{-5}$ per step. That is accurate relative to the training amplitudes but not below them. Its Jacobian at the origin matches the heat propagator to $0.1\%$ at the lowest wavenumbers but deviates from about $k=8$ onward, with spurious, slowly decaying gains near $k=16$ to $32$. The result is a floor in the rollout error, which stays between $4\%$ and $53\%$ as the amplitude falls through three decades instead of falling with it. For the anchored, centered lift the bias is at double-precision machine epsilon, the Jacobian equals the heat propagator to the same tolerance at every wavenumber, the nonlinear remainder scales as $I_s^{2.02}$, and the rollout error decreases without a floor. Centering without anchoring removes the offset but not the floor, which stays at $3\%$ at all small amplitudes, because its multi-step linearization is fitted rather than exact.

\end{document}